\RequirePackage{fix-cm}
\documentclass[smallextended]{svjour3}       
\smartqed  

\AtBeginDocument{%
  \setlength{\oddsidemargin}{\dimexpr(\paperwidth-\textwidth)/2-1in}%
  \setlength{\evensidemargin}{\oddsidemargin}%
  \setlength{\topmargin}{%
    \dimexpr(\paperheight-\textheight)/2-\headheight-\headsep-1in}%
}

\usepackage{graphicx}
\usepackage{mathptmx}      
\usepackage{latexsym}
\usepackage{amssymb}
\usepackage{array}

\usepackage{xcolor, colortbl}
\usepackage{multirow}
\usepackage[most]{tcolorbox}
\usepackage{booktabs}
\usepackage[caption=false, font=footnotesize]{subfig}
\usepackage{dblfloatfix}    

\usepackage[sort]{natbib}
\usepackage{url}

\journalname{Empirical Software Engineering}

\begin{document}

\title{Towards Understanding Quality Challenges of the Federated Learning for Neural Networks: A First Look from the Lens of Robustness}

\author{Amin Eslami Abyane \and Derui Zhu \and Roberto Souza\and Lei Ma \and Hadi Hemmati }


\institute{A. Eslami Abyane \at
              University of Calgary \\
              \email{amin.eslamiabyane@ucalgary.ca}
          \and
          D. Zhu \at
              Technical University of Munich \\
              \email{derui.zhu@tum.de}
          \and
          R. Souza \at
              University of Calgary \\
              \email{roberto.medeirosdeso@ucalgary.ca
          \and
          L. Ma \at
              University of Alberta \\
              \email{ma.lei@acm.org}
              \and
          H. Hemmati \at
              York University and University of Calgary \\
              \email{hemmati@yorku.ca} }
}

\date{Received: date / Accepted: date}
\titlerunning{Towards Understanding Quality Challenges of FL: A First Look from the Lens of Robustness}

\maketitle

\begin{abstract}
Federated learning (FL) is a distributed learning paradigm that preserves users' data privacy while leveraging the entire dataset of all participants. In FL, multiple models are trained independently on the clients and aggregated centrally to update a global model in an iterative process. Although this approach is excellent at preserving privacy, FL still suffers from quality issues such as attacks or byzantine faults. Recent attempts have been made to address such quality challenges on the robust aggregation techniques for FL.
However, the effectiveness of state-of-the-art (SOTA) robust FL techniques is still unclear and lacks a comprehensive study. Therefore, to better understand the current quality status and challenges of these SOTA FL techniques in the presence of attacks and faults, we perform a large-scale empirical study to investigate the SOTA FL's quality from multiple angles of attacks, simulated faults (via mutation operators), and aggregation (defense) methods.
In particular, we study FL's performance on the image classification tasks and use Deep Neural Networks as our model type.
Furthermore, we perform our study on two generic image datasets and one real-world federated medical image dataset. We also systematically investigate the effect of the proportion of affected clients and the dataset distribution factors on the robustness of FL.
After a large-scale analysis with 496 configurations, we find that most mutators on each user have a negligible effect on the final model in the generic datasets, and only one of them is effective in the medical dataset. Furthermore, we show that model poisoning attacks are more effective than data poisoning attacks. Moreover, choosing the most robust FL aggregator depends on the attacks and datasets. Finally, we illustrate that a simple ensemble of aggregators achieves a more robust solution than any single aggregator and is the best choice in 75\% of the cases. The data that support the findings of this study are available in our repository at \url{https://github.com/aminesi/federated}.

\end{abstract}

\section{Introduction}

Mobile devices have become quite powerful and essential part in our lives in recent years.  The increased computational power and the need and capability to support Deep Learning (DL) in many mainstream intelligent tasks such as image classification has helped introduce a new learning approach called Federated Learning (FL) \citep{mcmahan2017communicationefficient}. FL is a distributed learning paradigm where data resides only in each device of a user/client. Since data is only on clients' devices and is not transferred between clients and the server, FL can fully preserve clients' data privacy by design. As a result, FL is now being quickly adopted in privacy-sensitive areas like DL for medical imaging, where images are from different hospitals and patients' privacy is of utmost importance. With the recent trend of more strict data privacy regulation, FL continuously grows and is expected to expand its industrial adoption in the next few years.

In FL, clients receive a model from the server, perform the training on their data, and then send the new model to the server; then, the server aggregates the clients' models into a new model. This process continues multiple times till the training converges.

Like any other software, FL-based software systems may be prone to quality issues like robustness against adversarial attacks. Recent studies have identified several challenges with the FL process, including FL quality in the presence of attacks, heterogeneous data, and communication issues \citep{kairouz2021advances, flsurvey, flsurvey2}. One of the most studied topics within these quality challenges is the robustness of FL for byzantine attacks \citep{krumpaper, howtoback, localpoison, medianpaper, rsapaper, advlens, lyu2020privacy, backmodel}. Since clients can access the data and the training process, adversarial clients can cause various data and model poisoning attacks \citep{advlens}. So, many FL aggregation techniques have been proposed recently to make FL robust against these quality problems  \citep{krumpaper,medianpaper,rsapaper}.

Aside from being vulnerable to adversarial attacks, FL, like any other system, can also be prone to common faults. Mutation testing is a well-known technique in software testing that introduces minor faults (``mutation'') in the source code. It enables to assess the test cases' adequacy and sufficiency in detecting (killing) those faulty versions of the code (``mutants''). In recent years, mutation testing has also been applied to DL applications by defining mutation operators working on DL models and datasets \citep{munn,deepmu,8812047}. Some studies show the effectiveness of these mutants in terms of robustness analysis of DL programs for real faults \citep{deepcrime}.

To better understand the current state-of-the-art (SOTA) FL techniques against common attacks and quality issues, we perform experiments on four FL attacks and four DL mutation operators.
The attacks in our study are Label Flip \citep{localpoison}, Sign Flip \citep{krumpaper}, Random Update \citep{krumpaper}, and Backdoor attack \citep{backmodel}, and we use Noise, Overlap, Delete and Unbalance mutator to simulate faults \citep{deepcrime}. 
Since each attack/fault can appear independently in each client in FL, we further study the effect of the proportion of affected clients on the attack/fault's success.

Our study is focused on image classification tasks using Deep Neural Networks (DNNs) in FL. These choices are made because of the popularity of image classification tasks and DNNs in FL-related studies \citep{mcmahan2017communicationefficient, localpoison, medianpaper}. DNNs are also used because they are shown to outperform linear models in these tasks in most cases.

We also run the experiments on generic and real-world federated medical image datasets. Since generic datasets are not distributed, we distribute them between multiple clients to see the effect of distribution, and we make the distribution with three levels of non-iid.

We then evaluate four well-known aggregation techniques (Federated Averaging (baseline) \citep{mcmahan2017communicationefficient}, Krum \citep{krumpaper}, Median and Trimmed Mean \citep{medianpaper}). Finally, we study the feasibility of creating a more robust aggregator using the existing aggregation methods.

Our large-scale study consists of 496 configurations, each executed 10 times (i.e., 4,960 models trained and tested) to account for the randomness of algorithms. The results show that even the baseline FL aggregator (Federated Averaging) is quite robust against mutation operators on the generic image datasets. However, the Overlap mutator causes noticeable damage when applied to the medical dataset. Furthermore, our findings show that all examined attacks are effective against the baseline aggregator, but attacks that poison the model are more effective, as expected.

Our comparison of FL aggregators shows that Krum faces issues on more non-iid datasets and does not work as well as others, where all clients are benign.  We also observe that no single aggregation method can resist all the attacks (i.e., each has pros and cons). In other words, the FL aggregator's robustness depends on the dataset, the attack, and other factors like the data distribution. Finally, considering these strengths and weaknesses of aggregators, we propose a simple ensemble of existing aggregators. The results show that the ensemble aggregator can perform as well or even better than any of the aggregators alone in 75\% cases and achieves the highest accuracy on average.

To summarize, the contributions of this paper are:

\begin{itemize}
    \item Analysis of four FL aggregators under three untargeted attacks, a targeted attack, and four simulated data faults (using mutators) with different proportions of affected clients. 
    \item Analysis of different configurations of FL (cross-device and cross-silo) using both generic and medical datasets with different distributions.
    \item A simple ensemble aggregator that is a  better choice than any of its constituent aggregators in 75\% of the cases.
\end{itemize}

Given that FL is a promising ML technique for many privacy-aware use cases, this very early study in this direction has the potential to impact a broad set of DL software in practice. It also provides guidance on FL robustness areas that require further research.

\section{Background}
In this section, we briefly introduce basic concepts needed for understanding this paper, including FL and its attacks and defense mechanisms and mutation testing for deep learning.

\subsection{Federated Learning}
\label{background_fl}

\begin{figure}
    \centering
    \includegraphics[width=.9\linewidth]{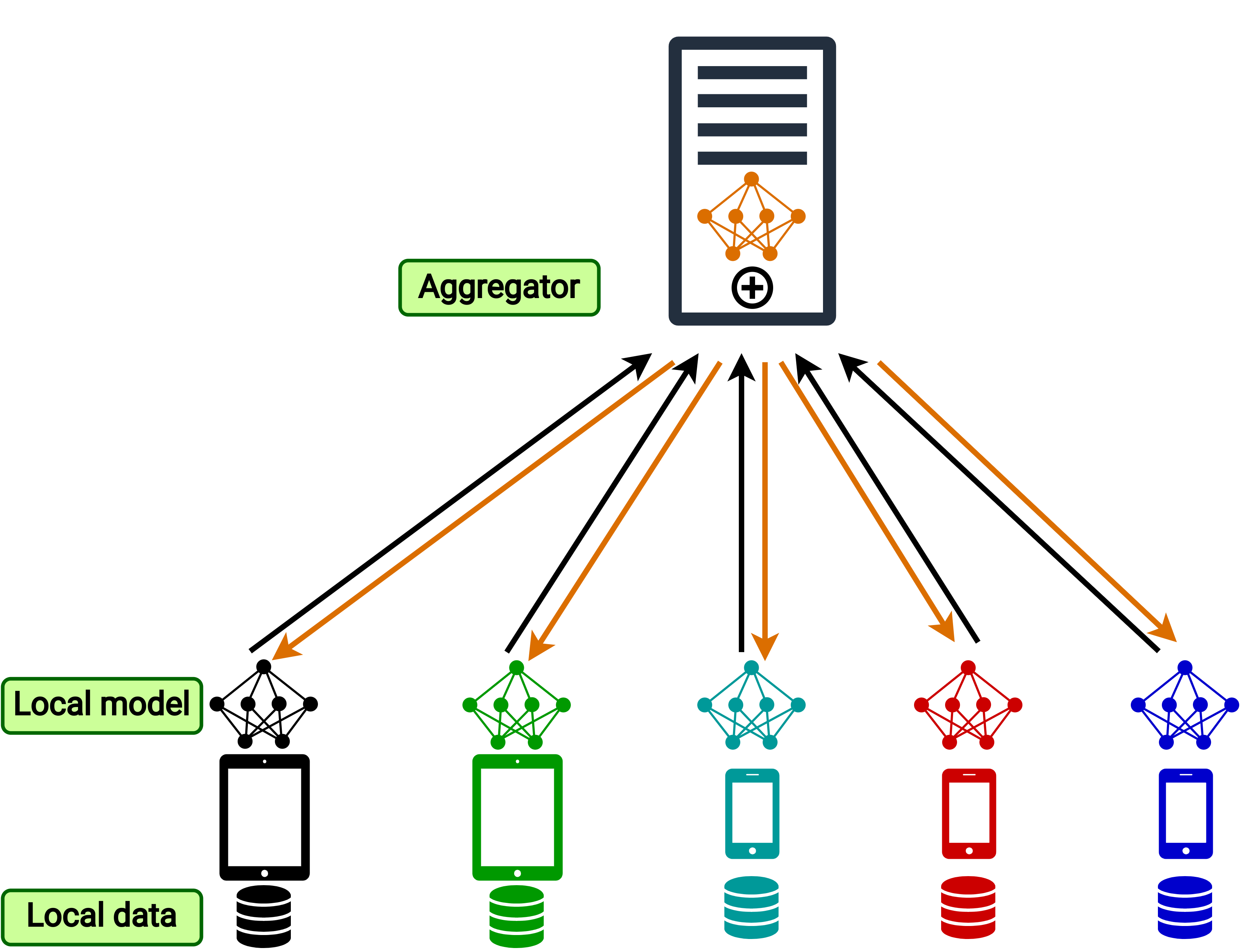}
    \caption{A typical Federated Learning workflow. The black arrows represent updates sent to the server, and the orange ones represent the updated global model sent to the clients. Each device has a dataset and a trained model which are represented by different colors.}
    \label{fl_flow}
\end{figure}

In recent years, a new learning technique has been proposed called Federated Learning \citep{mcmahan2017communicationefficient}. The goal of FL is to make learning possible on distributed data while maintaining clients' privacy. To that end, FL does not collect clients' data; instead, it performs learning on each client with their data. Figure \ref{fl_flow} shows how FL works along with the components we focus on in this study (green boxes).
At first, the server broadcasts the global model to the clients; then, each client trains their version of the model using their dataset for several epochs. When clients are done with their local training, they calculate the updates (the difference between the new model and the model received from the server) and send back the updates to the server. Once the server has all the updates, it aggregates them and updates the global model in the server. This process is a single round of communication, and FL requires multiple rounds of communication till the model converges.

Although FL initially focused on training on mobile devices, it now has a broader definition. FL that works on many devices is called cross-device FL, and if it works on a small number of clients (that are more reliable and accessible like centers), it is called cross-silo FL \citep{kairouz2021advances}.

In a cross-device setting, since not all the clients are available and training on all clients is extremely resource-consuming and has a significant communication overhead, only a fraction of clients is selected at each round. Selecting a small fraction can hurt convergence speed, but \citet{mcmahan2017communicationefficient} have shown that a small fraction, such as 10\% of clients, can produce good results. In contrast, in cross-silo FL, all the clients are available, and training is done on all of them. Moreover, since the number of clients is small, communication overhead is often not an issue.

In this paper, we will study how the attacks/faults and defenses can impact the quality of the FL process. To that end, we need to alter three components of the basic FL procedure: aggregator, local model (update), and local dataset, shown as green boxes in Figure \ref{fl_flow}. The first component is where we implement the defense techniques, and the other two are where attacks and faults are implemented.

\subsection{Aggregation methods in Federated Learning}
\label{background_aggregation}

As discussed, FL aggregates all updates to update the model on the server. There have been many aggregation techniques proposed for FL with different characteristics. We discuss the most important and well-known ones here.

\textbf{Federated Averaging} \citep{mcmahan2017communicationefficient}:
This technique is the first, most straightforward, and perhaps the most well-known among all aggregation methods. The process of Federated Averaging is relatively straightforward. When the aggregator receives all the client updates, it calculates the mean of all the values for each axis and generates the aggregated model. Federated Averaging has shown to be effective in FL under different data distributions. However, it has a big flaw: it cannot perform well when adversaries operate some clients. That is why new approaches have been proposed under the robust aggregation category, which we will discuss next.
This aggregator is the baseline in all of our experiments as it is the most used non-robust aggregator in related studies 
 \citep{foolsgold, krumpaper, sear}. Other variations of Federated Averaging \citep{fedma, fedprox} aim to improve Federated Averaging in non-iid scenarios but are still not designed as robust aggregators. Thus we still use Federated Averaging as the baseline following previous studies.

\textbf{Krum} \citep{krumpaper}: This is one of the first robust aggregation techniques. The main idea here is that clients that produce similar updates are most likely benign, and choosing any of them as the update is reasonable for the global model update. More specifically, it first calculates the distances between clients' updates, then sorts the clients based on the sum of their distance to their closest k clients. Where \(k = n - f - 2\), n is the number of clients, and f is the number of byzantine clients. Finally, it picks the client with the lowest sum as the update for the model. Since it chooses the update of only one client, it causes some concerns for privacy and its performance under non-iid distributions.

\textbf{Median} \citep{medianpaper}: Coordinate wise Median is another robust aggregation method. As the name suggests, Median calculates the median of clients' updates per coordinate (if updates are like arrays, it gets the median for each axis and index) and creates an aggregated update. Intuitively it tries to filter outlier values.

\textbf{Trimmed Mean} \citep{medianpaper}:
This approach is very much like the Federated Averaging. However, instead of getting the mean over all client updates, it first excludes a fraction of the lowest and highest values per coordinate and then calculates the mean.

\subsection{Attacks in Federated Learning}
\label{background_attack}

Byzantine faults are known issues in distributed settings 
 \citep{byzantine}. Since FL is a distributed technique, it faces byzantine faults and attacks that can cause quality problems. In byzantine attacks, clients collude to achieve goals like breaking the learning process. Byzantine attacks are not to be mistaken with adversarial attacks applied in the testing phase, which try to fool the model by adding perturbations to the data \citep{fgsm,pgd}.
Moreover, clients have access to both data and the training process (local training), so they can cause problems for the entire training process if adversaries operate them. A recent study has categorized FL attacks into two groups: data poisoning and model poisoning \citep{advlens}.
The first group is not exclusive to FL, and it has been a problem in centralized training as well  \citep{10.1145/3128572.3140451}. However, since clients have access to the model and training process, FL must withstand a new category of model poisoning attacks. In model poisoning, adversaries poison the model updates before sending them to the server and attack the learning process.

More generally, attacks can be categorized into two groups, namely untargeted and targeted \citep{lyu2020privacy}. In untargeted attacks, adversaries want to stop the model from achieving its goal. For instance, the attacker's goal in the classification task is to make the model misclassify. However, the attacker has a specific goal in a targeted attack, like misclassifying images into a particular label in image classification.
In the following, we introduce some of the most important attacks in FL from both categories since our study aims to investigate how current SOTA FL methods perform against different attacks and whether quality issues would occur.
These attacks are significant due to their success rate and utilization in related works \citep{localpoison, advlens, krumpaper, rsapaper, howtoback}.

\textbf{Label Flip}: This is an example of a data poisoning attack \citep{localpoison} in which byzantine clients change labels of their dataset randomly to make the final model inaccurate. Since the changed label is selected based on a uniform distribution, it is considered an untargeted attack. According to \citet{advlens}, this is the most effective attack in the FL setting among data poisoning attacks.

\textbf{Random Update}: This is a model poisoning attack where the adversary does not send an update based on its training. Instead, it sends random Gaussian updates to the server  \citep{krumpaper,localpoison} which also makes it untargeted.
Since the client is sending the update, it can increase the distribution variance to make more powerful attacks. This was impossible in data poisoning attacks (the number of labels that can be altered is limited).

\textbf{Sign Flip attack}: Another example of a model poisoning attack is Sign Flip, where the attacker trains the model on its dataset and calculates the updates, but it changes the sign of updates and sends an update in the opposite direction  \citep{krumpaper,rsapaper}. The client can also multiply the update to make the attack more practical, like the Random Update. This attack is also untargeted as the attacker does not pursue a particular goal, but it is more directed than the Random Update attack.

\textbf{Backdoor attack}: In contrast to all previous attacks, this attack is targeted and tries to achieve a specific goal. The Backdoor attack aims to make the model misclassify images containing certain features as a specific class  \citep{backdoordata}. The Backdoor attack can be categorized into data or model groups based on its implementation. Papers that are not focused on FL, only use data to add a backdoor to the model \citep{backdoordata,badnet}. However, FL papers have used both data and model updates to make the attacks more effective \citep{howtoback,backmodel}. \citet{howtoback} have shown that Backdoor can be done using a semantic feature like classifying green cars as birds. The semantic feature has a significant advantage over other techniques: it only requires a training time attack, and data does not have to be changed in testing time. Although this makes the semantic feature a great option, this approach is not easily generalizable for different datasets.

\begin{figure*}
    \centering
    \subfloat[Original image\label{backdoor_original}]{%
\includegraphics[width=0.2\linewidth]{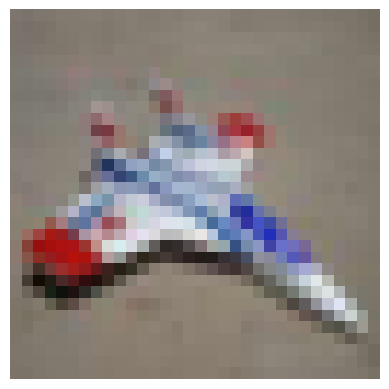}}
    \hspace{0.05\linewidth}
    \subfloat[Image with backdoor pixel pattern\label{backdoor_pixel}]{%
\includegraphics[width=0.2\linewidth]{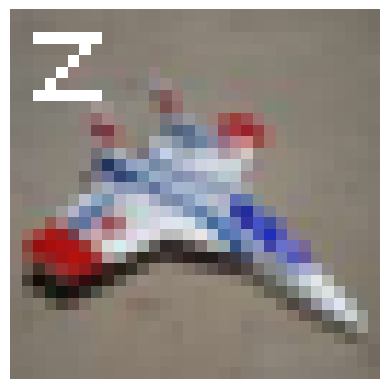}}\hfill

    \caption{An example of the Backdoor attack (with a pixel pattern) on a sample from the CIFAR-10 dataset that will force the classifier to misclassify a plane as a car.}
    \label{backdoor_fig}
\end{figure*}

Another Backdoor technique is called pixel pattern, which was introduced in \citep{badnet}. This technique works by adding a small pattern of pixels to the images, which acts as the backdoor. This approach requires a test time attack and a training time attack, but it is more scalable than semantic features.
Figure \ref{backdoor_fig} shows an example of this type of Backdoor attack. In this example, the "Z" pattern added to the top left corner of the plane image will make the model misclassify the image (as a car, for instance). 

Finally, byzantine clients multiply their updates and make their updates dominant in the aggregation phase to make a more powerful attack. This multiplication is more effective if the model has converged since other updates are small, and byzantine updates can replace the global model. In the Backdoor attack, there should be enough benign samples in the training batch. Otherwise model classifies all samples as the target class, and the main task fails, which is not desirable for the attacker.

\subsection{Mutation operators in Deep Learning}
\label{background_mutation}
Mutation testing is a well-known topic in the Software Engineering community, and recently it has been applied in DL  \citep{deepmu,munn}. Even though these studies have introduced many mutation operators for DL, many of these operators do not reflect real-world faults. More recently, a new mutation operator set has been proposed that is based on real faults  \citep{deepcrime}. These operators consist of the following categories: training data, hyper-parameters, activation function, regularisation, weights, loss function, optimization function, and validation.

In this paper, we leverage mutation testing to simulate the potential quality issues and how they impact the quality of SOTA FL techniques.
In particular, we focus on mutation operators from the data category to simulate actual (potential) faults in the federated context. Other categories are mostly related to the training process, and although clients have access to that, if faults were to happen, they would be happening on all clients as the code for all clients is the same. Thus, it is unlikely that clients can change the model parameters by mistake. As a result, other categories are not applicable in the FL setting (they make more sense in centralized learning).
An important note is that the selected mutators are not specific to FL; rather, they apply to FL. Since there are no FL-specific mutators at the time of conducting this study, we use these FL-applicable mutators.

Now we discuss mutation operators related to training data.

\textbf{Change labels of training data}: This is like a Label Flip attack, so we will not discuss it more.

\textbf{Delete mutator}: This operator simulates the lack of training data by removing a portion (we call this delete percentage in our experiments) of training data.

\textbf{Unbalance mutator}: This operator simulates the situation where data is unbalanced by removing a portion (we call this unbalance percentage in our experiments) of samples in classes that occur less than average.

\textbf{Overlap mutator}: This operator simulates a situation where very similar samples have different classes. It works by finding the two most dominant classes, then copies a portion (we call this overlap percentage in our experiments) of data from one class and labels it as the other.

\textbf{Noise mutator}: This imitates a scenario where training data is noisy by adding Gaussian noise to the samples. This mutator takes the variance of the pixels in the image. Then, it adds noise to the image with a mean of zero and a multiply (we call this sigma multiplier in our experiments) of calculated variance.
This mutator can represent a category of faults where the clients' images are occluded because of the problems with the camera, for instance.

\section{Empirical study}
\subsection{Objectives and research questions}

The main objective of this paper is to investigate how current SOTA FL performs under different potential quality contexts, such as attacks and faults, using regular and robust aggregators. To achieve this, we mainly consider the following research questions:

\textbf{RQ1: How robust is Federated Averaging against attacks and faults in a well-known image domain dataset?}

In this RQ, we focus on the robustness of the Federated Averaging as the baseline aggregator in FL. We study four attacks and four mutators and consider other factors such as data distribution and the proportion of affected clients to comprehensively study the attack/fault effects on the final results. In this RQ, we use well-known image datasets where we can have control over these factors.
The FL setting for this RQ and RQ2 is cross-device.

\textbf{RQ2: How effective are the existing robust aggregation techniques against these attacks and faults?}

Following the same settings as RQ1, in this RQ, we study FL aggregators that are built as a robust technique along with Federated Averaging and evaluate their robustness against different attacks.

\textbf{RQ3: In a real-world federated dataset, what is the effect of attacks and faults on different aggregators?}

This RQ aims to evaluate the aggregation techniques in a cross-silo federated setting where clients are different centers. Note that we do experiments similar to RQ1 and RQ2 just on a real Federated dataset and a cross-silo setting. So in between RQ1 and RQ3, all the experiments are done on both cross-device and cross-silo settings. We will go through more details in the design section.

\textbf{RQ4: How robust is an ensemble method that combines existing aggregators in detecting all attacks and mutations in all datasets and configurations?}

This RQ aims to investigate the possibility of having a general solution that works well for all configurations of untargeted attacks without knowing what attack or fault the system will be facing.

\subsection{Experiment design}

\begin{figure}
    \centering
    \includegraphics[width=1\linewidth]{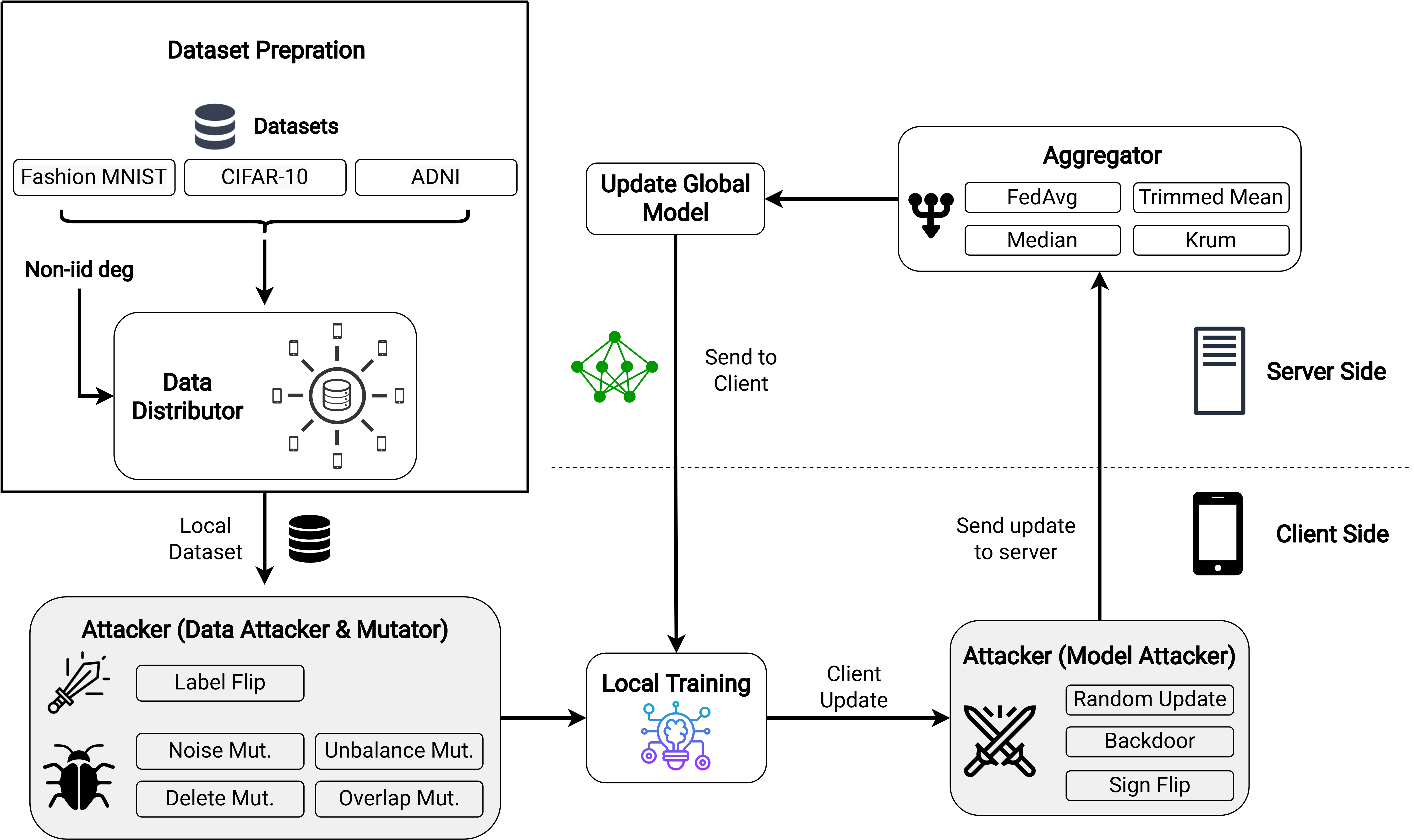}
    \caption{Overview of our experiment design. (Only one of the attacker components, highlighted with grey color, is used in our experiments at a time)}
    \label{overview}
\end{figure}

This section describes the design of our empirical study. Figure \ref{overview} shows an overview of our experiments procedure. 
\subsubsection{Datasets and models}
Image classification is one of the leading DL tasks, and it is being used as a primary subject in many studies in FL  \citep{mcmahan2017communicationefficient, krumpaper}. Thus, we choose this task as our main task.

\textbf{Generic datasets}:
For the first two questions, which are focused on the generic image dataset, we choose two popular datasets from the image classification task, Fashion MNIST \citep{fashion}, and CIFAR-10 \citep{cifar}.
The CIFAR-10 dataset consists of 60,000 32x32 color images in 10 classes, with 6000 images per class. The images are split into 50000 training images and 10000 test images.
Fashion MNIST consists of a training set of 60,000 samples and a test set of 10,000 samples each being a 28x28 gray-scale image. There are a total of 10 classes in this dataset.

We use these datasets since they are well-known and well-studied and represent different classification difficulty levels.
Lastly, in RQ1 and RQ2, we study the effects of different datasets and models on the quality of the FL process when it is under byzantine attacks and mutation faults. We also chose these centralized datasets to control how we distribute them and study the effect of data distribution.

\textbf{Federated dataset}: 
In RQ3, we aim to study a more realistic scenario and see its effect on FL quality. Since tasks on medical images are one of the main FL applications \citep{kairouz2021advances}, given patients' privacy concerns, a distributed medical image dataset is a perfect match for RQ3's goal. Our medical imaging dataset is obtained from the Alzheimer's Disease Neuroimaging Initiative (ADNI) database \citep{adni}. The ADNI was launched in 2003 as a public-private partnership led by Principal Investigator Michael W. Weiner, MD. The primary goal of ADNI has been to assess and model the progression of mild cognitive impairment (MCI) and early Alzheimer's disease (AD). For up-to-date information, see \url{www.adni-info.org}. In this work, we are leveraging 1,723 magnetic resonance images from  594 ($34.5\%$) Alzheimer's patients and
1,129 ($65.5\%$) presumed normal controls. These data were collected from six different centers that are used to simulate our federation. The number of data samples is unbalanced across centers (A: $15.7\%$, B:$16.7\%$, C:$4.3\%$, D:$12.5\%$, E:$14.9\%$, F:$35.9\%$). We extract the 10 central slices in the axial plane of the images to train and test our models.

\textbf{Data distribution}:
There is no need for synthetic distribution for the ADNI dataset since it is already distributed. However, since CIFAR-10 and Fashion MNIST datasets are centralized, we partition them and distribute samples to different clients. Furthermore, because we want to see the effect of the distribution of the dataset in RQ1 and RQ2, we partition these datasets with different degrees of non-iid.
We include both iid (uniformly distributed between clients), and non-iid distributions since an iid distribution represents an ideal case. However, in real-world FL scenarios, the data tends to be more non-iid  \citep{mcmahan2017communicationefficient}.

To simulate different iid distributions, we use a method similar to the method used in the original FL paper  \citep{mcmahan2017communicationefficient}, but we add multiple non-iid options. We first group the images based on the classes and then split each group's samples into chunks of equal size (this size increases with the non-iid parameter and depends on the number of clients).
Then for each client, we select a subset of groups randomly, and from each group, we select a chunk randomly and assign it to that client.
This process finishes when all chunks are assigned to the clients.
The number of selected groups for each client depends on the non-iid degree, and with the increase of non-iid degree, the number of selected groups decreases. However, all clients receive the same amount of samples as chunks are larger for higher non-iid degrees.

For instance, all groups will be selected with a non-iid degree of 0, and clients will get data from all classes. However, in a high non-iid scenario, only a couple of groups will be selected, and each client will have samples from only some classes.

\begin{table}
\setlength\tabcolsep{3pt}
    \centering
    \subfloat[Fashion MNIST\label{mnist_model}]{
    \begin{tabular}{p{0.16\linewidth}ll} 
\toprule 
 Layer                     &           Output Shape             &                Param \#        \\ \midrule 
 conv2d\_0                    &           (None, 26, 26, 32)       &                320            \\ 
 max\_pooling2d\_0     &           (None, 13, 13, 32)       &                0              \\ 
 conv2d\_1                  &           (None, 11, 11, 64)       &                18496          \\ 
 max\_pooling2d\_1     &           (None, 5, 5, 64)         &                0              \\ 
 flatten\_0                  &           (None, 1600)             &                0              \\ 
 dropout\_0                  &           (None, 1600)             &                0              \\ 
 dense\_0                      &           (None, 10)               &                16010          \\ \midrule 
\makebox[0pt][l]{Total params: 34,826} \\ 
\makebox[0pt][l]{Trainable params: 34,826} \\ 
\makebox[0pt][l]{Non-trainable params: 0} \\ \bottomrule 
\end{tabular}
    }
    \hfill
    \subfloat[CIFAR-10\label{cifar_model}]{%
        \begin{tabular}{p{0.16\linewidth}ll} 
\toprule 
 Layer                       &           Output Shape             &                Param \#        \\ \midrule 
 conv2d\_0                    &           (None, 32, 32, 32)       &                896            \\ 
 conv2d\_1                  &           (None, 30, 30, 32)       &                9248           \\ 
 max\_pooling2d\_0       &           (None, 15, 15, 32)       &                0              \\ 
 dropout\_0                  &           (None, 15, 15, 32)       &                0              \\ 
 conv2d\_2                  &           (None, 15, 15, 64)       &                18496          \\ 
 conv2d\_3                  &           (None, 13, 13, 64)       &                36928          \\ 
 max\_pooling2d\_1     &           (None, 6, 6, 64)         &                0              \\ 
 dropout\_1                &           (None, 6, 6, 64)         &                0              \\ 
 flatten\_0                  &           (None, 2304)             &                0              \\ 
 dense\_0                      &           (None, 512)              &                1180160        \\ 
 dropout\_2                &           (None, 512)              &                0              \\ 
 dense\_1                    &           (None, 10)               &                5130           \\ \midrule 
\makebox[0pt][l]{Total params: 1,250,858} \\ 
\makebox[0pt][l]{Trainable params: 1,250,858} \\ 
\makebox[0pt][l]{Non-trainable params: 0} \\ \bottomrule 
\end{tabular}
        }
    \caption{Model summary for generic datasets.}
    \label{model_summary}
\end{table}

\textbf{Models}:
Our models for generic datasets are simple convolutional neural networks with 12 and 7 layers for CIFAR-10 and Fashion MNIST, respectively, which are taken from Keras' tutorials and examples \citep{cifarmodel,mnistmodel}. Table \ref{model_summary} shows a summary of the models used for Fashion MNIST and CIFAR-10 datasets.
For the ADNI dataset, we use a transfer learning (TL) approach using the VGG16 model \citep{vgg16} pre-trained on ImageNet  \citep{imagenet}. Our classifier consists of one dense layer with 512 neurons and a rectified linear activation followed by a dense layer with two neurons and a softmax activation. The VGG16 weights were frozen during training.

An important note is that these model architectures have many different parameters that can be tuned to ensure higher robustness against attacks. However, our study focuses on aggregator techniques in FL to improve the overall robustness. Consequently, we use the models exactly as they were proposed.
Lastly, Our selected models for the experiments ensure the variety in our cases and help our results be generalizable.

\subsubsection{Attacks and faults}
We use attacks described in Section \ref{background_attack} namely: Label Flip attack, Random Update attack, Sign Flip attack, and Backdoor attack.
Our choice of attacks contains both untargeted and targeted attacks. Moreover, we have attacks from both data poisoning and model poisoning categories, which again helps with the generalizability of our study.
We set the mean and standard deviation of Gaussian distribution for Random Update zero and two, respectively. We set the multiplier for sign attack to 10. We choose the pixel pattern technique discussed before for the Backdoor attack as it makes it generalizable for different datasets, and we set the update multiplier to 10. These numbers were selected based on empirical evaluations and how they were used in related works to ensure the attacks' effectiveness \citep{howtoback, krumpaper}.

To simulate faults, we choose all data-related mutations discussed in Section \ref{background_mutation}. We set the delete percentage, unbalance percentage, and overlap percentage discussed in the mutations background section to 75\%. Also, we set the sigma multiplier for the Noise mutator to one. Since these mutators were not tested on FL before, we show the reason behind these choices in the RQ1 section.
Moreover, since we are investigating  FL-related (by this term, we mean faults that can happen in FL and are not exclusive to FL) faults, we need to consider what faults can be caused by clients' mistakes. The choice of model-related parameters and mutation operators makes it more of a centralized choice, not a client choice.
Note that although the faults we cover are not exclusive to FL, they apply to FL since FL is a federated version of a regular learning system, so analyzing their impact in FL is important to see how they can affect the overall system quality, just like a normal learning system.

Finally, to thoroughly investigate how much damage attacks and faults can cause in FL, we choose three different proportions of affected clients: 0.1, 0.3, and 0.5, and study the effect on the final model.

\subsubsection{Aggregation methods}

As discussed in Section \ref{background_aggregation}, we select Federated Averaging as our baseline method and Krum, Median, and Trimmed Mean as the robust aggregation methods under study, to represent the most well-known aggregation methods from the literature. We set the hyperparameters for Krum and Trimmed Mean based on the expected number of malicious clients.

To show the feasibility of an ensemble aggregator, in RQ4, we choose two attacks, Label Flip and Sign Flip, to have both data and model poisoning attacks (we discuss these choices more in the results section of RQ4). Our ensemble aggregator performs the aggregation with all four mentioned aggregators, then picks the aggregated update with the best validation accuracy. This process happens at each round independently until the training is complete. A note about this approach is that the ensemble does not create a weighted average of all the aggregators. Rather, it selects the result of the best one. Thus, it does not require weight tuning.  

\subsubsection{Federated setup}

For the RQ1 and RQ2, we follow federated settings in  \citep{mcmahan2017communicationefficient}. We distribute data to 100 clients and select 10 of them randomly each round to simulate a cross-device setting. Local epoch and batch size are set to 5 and 10, respectively. Furthermore, we repeat experiments for each dataset with three levels of non-iid: 0, 0.4, and 0.7. The number of training rounds depends on the dataset and how fast the model can converge. For Fashion MNIST, it is set to 100, and for CIFAR-10, it is set to 1,000.

For the RQ3, since we have six centers in the ADNI dataset and a cross-silo setting, we select all clients at each round for training. Moreover, we set the batch size to 32, the local epoch is set to one, and the number of training rounds is 100.

In RQ4, we conduct experiments on the CIFAR-10 (with the non-iid degree of 0.4) and ADNI datasets to compare our proposed aggregator with existing ones.  

Moreover, in all of the comparisons, we compare the accuracies and also run non-parametric statistical significance tests (Mann–Whitney U test with p\_value less than 0.05) on the 10 runs of each configuration per aggregator to show that the results are not due to chance.

Finally, in all RQs, to eliminate the randomness introduced in different steps of the experiment, we run all configurations 10 times and always report the \textit{median} of 10 runs. So all accuracy values in figures are median accuracies. The reason behind this choice is that the median is better at filtering the outlier values that might happen in the 10 runs, which will result in more consistent results.

Furthermore, the mean and std values reported in the tables are taken across different configurations (e.g., non-iid degree) described in their related sections. The mean is taken over different configurations instead of the median since we want to consider all configurations equally in the final result. These mean and std values should not be confused with the median taken across different runs (mean and std are applied to the results of different configurations reported in different figures.).  Also, other metrics like the median of different configurations and also the raw results for each configuration can be found in our replication package.

\subsubsection{Threat model}
\label{threat_model}
We assume the attacker has complete control of a proportion of clients; thus, they can alter their model updates sent to the server and poison the training data. Furthermore, the devices can collude to make attacks more powerful like \citep{howtoback, sear, krumpaper}. However, the attacker does not know the server's aggregation method, which is the same assumption used by previous works \citep{krumpaper, localpoison}.

In our opinion, some aggregators like Krum need to know the number of attacked clients, which is an unrealistic assumption. However, given that it has been used extensively before,  \citep{ditto, localpoison}, we also include this configuration in our study.
It might be possible to make Krum work without knowing the exact number of attackers (using an estimation approach), but that is not the focus of our study. 

Lastly, the server and the aggregation techniques are assumed to be uncompromised following previous studies' settings  \citep{foolsgold}.

\subsubsection{Execution setup and environment}


All experiments are done on Compute Canada cluster nodes with 4 CPU (Intel Gold 6148 Skylake @ 2.4 GHz) cores, a GPU (NVidia V100SXM2 with 16GB memory and 125 TFLOPS computation power), and 64GB of RAM running on CentOS 7. We use TensorFlow 2.4  \citep{tensorflow2015-whitepaper} as our DL framework and Python 3.8 and simulate FL on a single machine.

\subsubsection{Evaluation metrics}

We use the model's prediction accuracy on test split as our metric for all attacks and mutators except the Backdoor attack.
In the Backdoor attack, we use the accuracy of the backdoor task as the metric and not the accuracy of the main task. The reason is that the backdoor attack is trying to fool the model on the backdoor task and not the main task so the main task accuracy is virtually the same as a clean scenario. Furthermore, we choose this metric to be consistent with related works that study the backdoor task \citep{howtoback, advlens, backmodel}. In contrast to other attacks, the Backdoor attack's effectiveness directly correlates with this metric.

Finally, whenever we mention accuracy in the rest of the study, we mean prediction accuracy. Also, when we want to talk about the accuracy of the backdoor task, we use the term backdoor accuracy.

\subsection{Experiment results}

In this section, we discuss the results for RQ1 to RQ4.

To avoid confusion, we first look at the effect of faults (mutators) in all sections, then we discuss the attacks on FL.

\subsubsection{\textbf{RQ1 results (effect of attacks and faults on Federated Averaging):}}
\label{results_rq1}

The mutators, as discussed in \ref{background_mutation} have different parameters that can determine how effective they can be. To ensure we have chosen reasonable parameters, we do some preliminary experiments on CIFAR-10 with the non-iid level of 0.4 while the proportion of the affected clients is set to 0.3.
We test three different values for the mutators that alter a percentage of the data: 25\%, 50\%, and 75\%. We choose 0.1, 0.5, and 1 as the candidates for the Noise mutator's sigma multiplier.

\begin{figure*}
    \centering
    \subfloat[Delete Mutator\label{params_delete}]{%
        \includegraphics[width=0.40\linewidth]{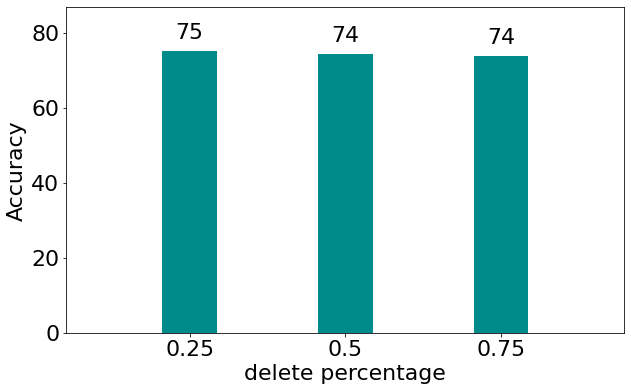}}
    \hspace{0.05\linewidth}
    \subfloat[Unbalance Mutator\label{params_unbalance}]{%
        \includegraphics[width=0.40\linewidth]{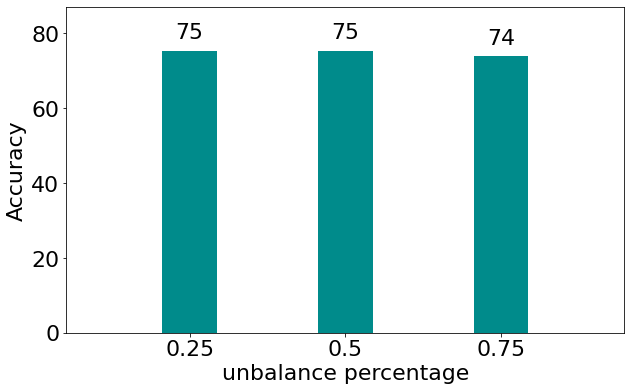}}
    \hfill
    \subfloat[Overlap Mutator\label{params_overlap}]{%
        \includegraphics[width=0.40\linewidth]{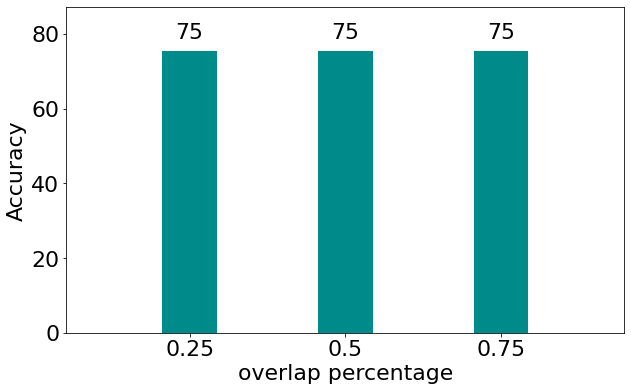}}
    \hspace{0.05\linewidth}
    \subfloat[Noise Mutator\label{params_noise}]{%
        \includegraphics[width=0.40\linewidth]{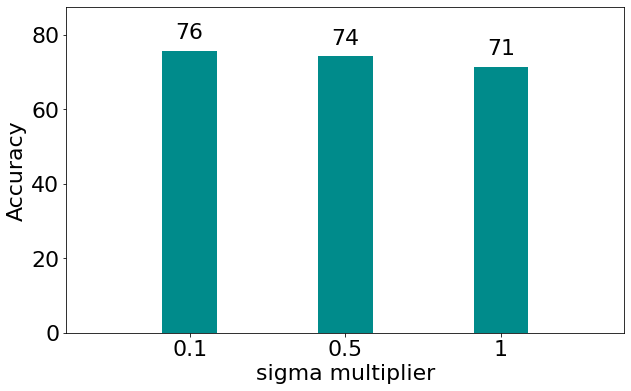}}    
    \caption{Effect of mutator parameters on CIFAR-10 datset with non-iid level of 0.4 and affected clients proportion of 0.3.}
    \label{params_fig}
\end{figure*}

The results are reported in Figure \ref{params_fig}.
As shown in Figures \ref{params_delete} to \ref{params_overlap}, increasing the percentage parameter of Delete, Unbalance, and Overlap mutators does not have a significant impact on the final accuracy. However, in all cases, 75\% gives the best results. So for these mutators, we set the percentage parameter to 75\%.

Nevertheless, increasing the sigma multiplier seems to have a more noticeable effect on the Noise mutator, as shown in \ref{params_noise}. Given these results, we set the sigma multiplier to be one in the remainder of our experiments. 

However, unlike the other mutators that could have a max percentage of 100\%, the sigma multiplier does not have a max sealing here. So a valid question might be why we do not set the multiplier even higher? 

\begin{figure*}
    \centering
    \subfloat[Original image\label{sigma_0}]{%
\includegraphics[width=0.2\linewidth]{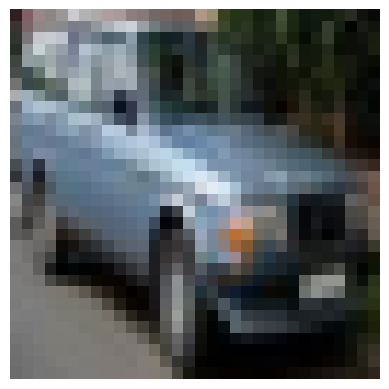}}
    \hspace{0.05\linewidth}
    \subfloat[multiplier=0.1\label{sigma_0.1}]{%
\includegraphics[width=0.2\linewidth]{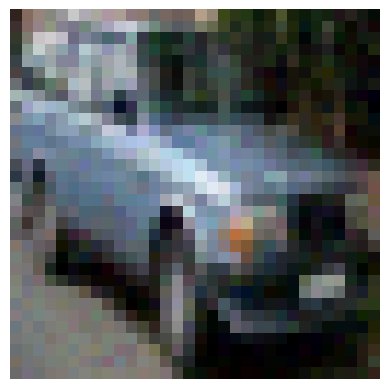}}
    \hspace{0.05\linewidth}
    \subfloat[multiplier=0.5\label{sigma_0.5}]{%
\includegraphics[width=0.2\linewidth]{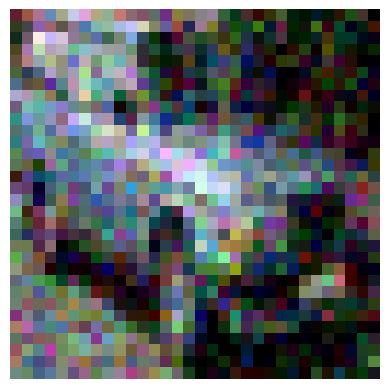}}
    \hspace{0.05\linewidth}
    \subfloat[multiplier=1\label{sigma_1}]{%
\includegraphics[width=0.2\linewidth]{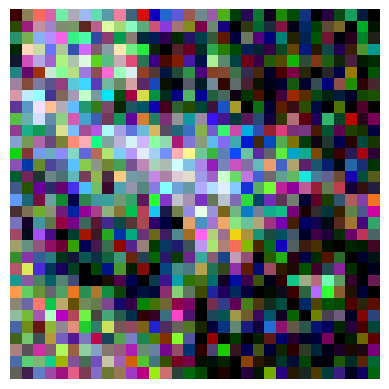}}
    \hspace{0.05\linewidth}
    \subfloat[multiplier=2\label{sigma_2}]{%
\includegraphics[width=0.2\linewidth]{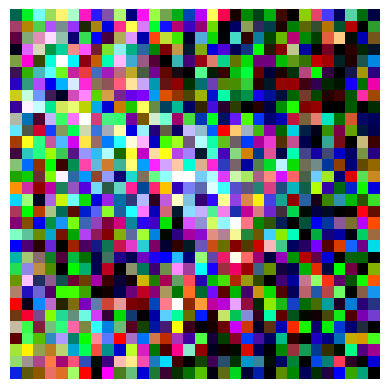}}
\hfill

    \caption{Effect of the Noise mutator's sigma multiplier on a sample from CIFAR-10 datset.}
    \label{sigma_fig}
\end{figure*}

To address this concern, we show how the image is affected when it gets mutated by the Noise mutator with different sigma multipliers in Figure \ref{sigma_fig}. As the sigma multiplier increases, the image gets noisier as expected. However, if we increase it too much, the image gets unrecognizable, like in Figure \ref{sigma_2}. Since the Noise mutator is supposed to simulate faults, extreme values for the sigma multiplier become unacceptable, and this is the reason that we set this parameter to be one in our experiments.

\textbf{Federated Averaging's performance under simulated faults:}

\begin{figure*}
    \centering
    \subfloat[Delete Mutator\label{cifar_fedavg_delete}]{%
        \includegraphics[width=0.40\linewidth]{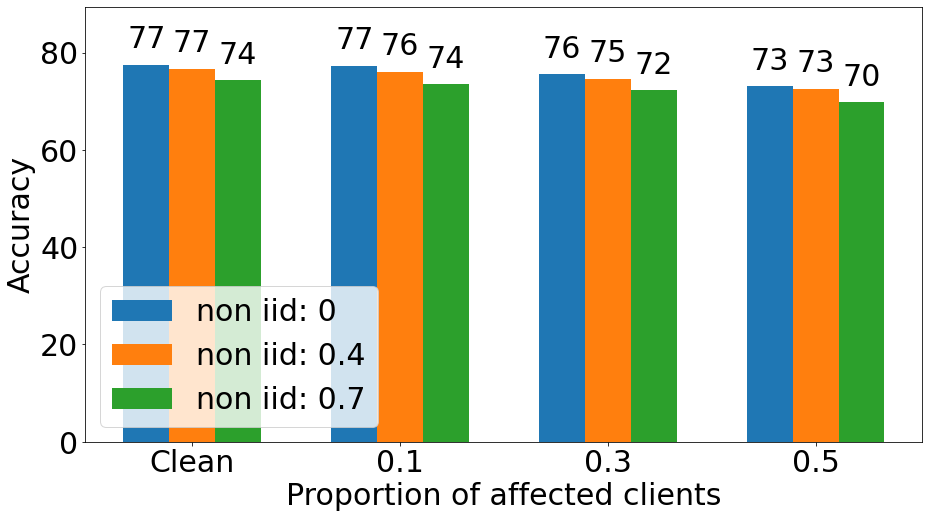}}
    \hspace{0.05\linewidth}
    \subfloat[Unbalance Mutator\label{cifar_fedavg_unbalance}]{%
        \includegraphics[width=0.40\linewidth]{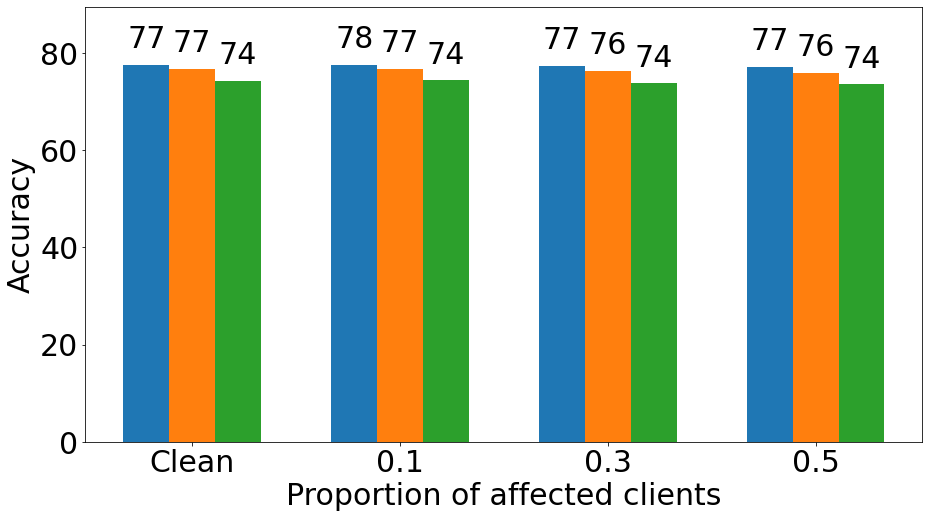}}
    \hfill
    \subfloat[Overlap Mutator\label{cifar_fedavg_overlap}]{%
        \includegraphics[width=0.40\linewidth]{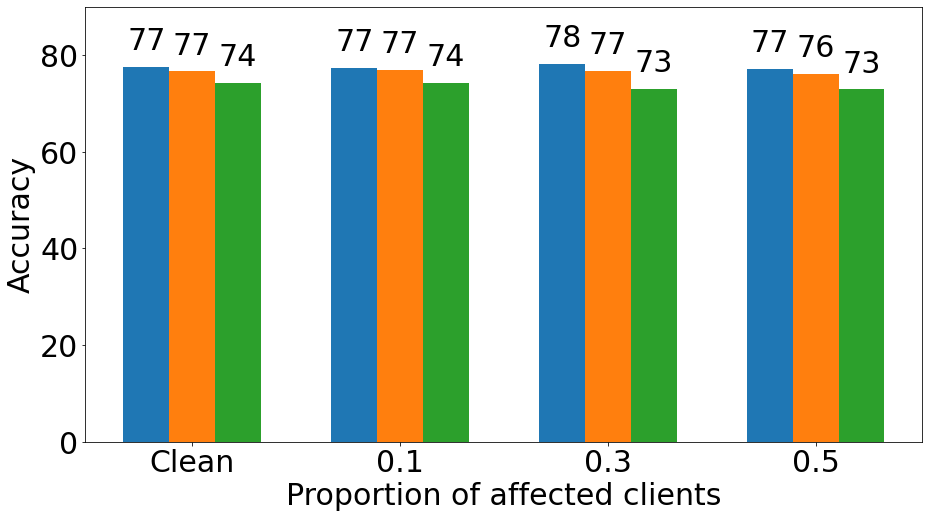}}
    \hspace{0.05\linewidth}
    \subfloat[Noise Mutator\label{cifar_fedavg_noise}]{%
        \includegraphics[width=0.40\linewidth]{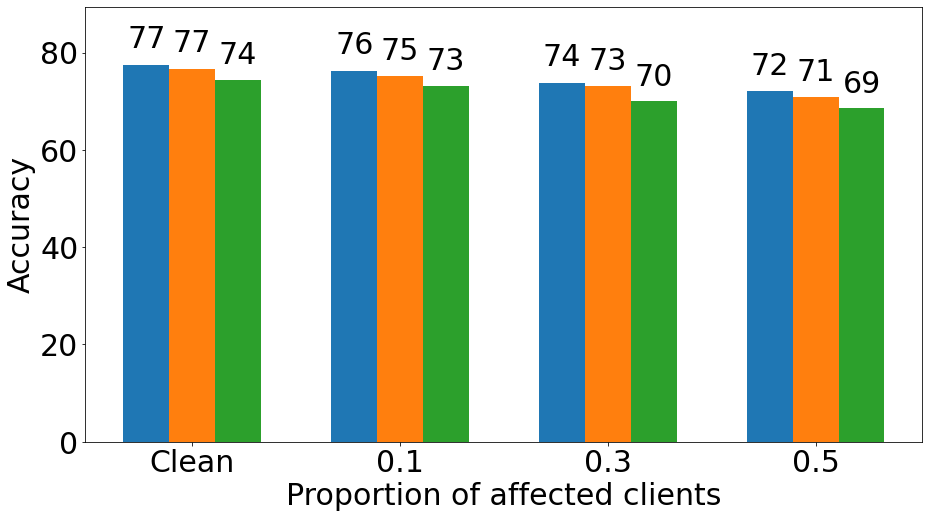}}
    \hfill
    \caption{CIFAR-10 - Federated Averaging performance under simulated faults.}
    \label{cifar_fedavg_faults_fig}
\end{figure*}

\begin{figure*}
    \centering
    \subfloat[Delete Mutator\label{mnist_fedavg_delete}]{%
        \includegraphics[width=0.40\linewidth]{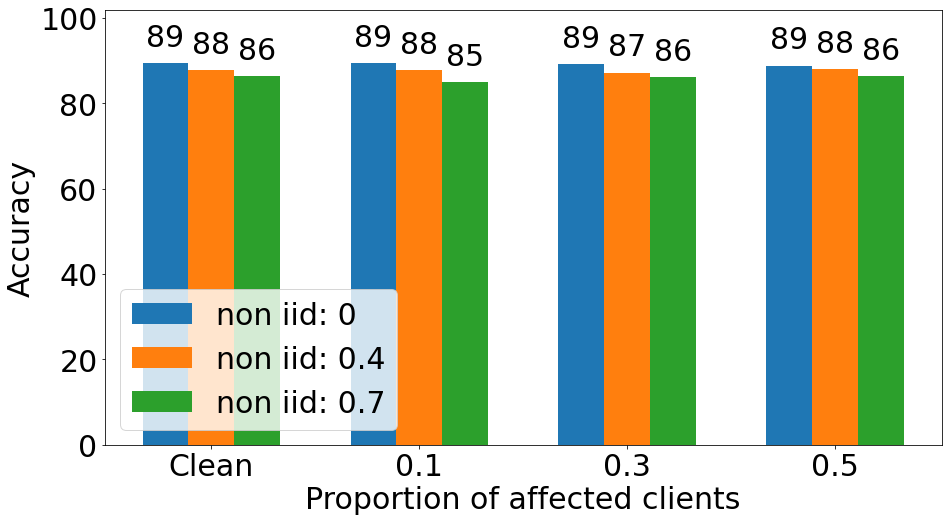}}
    \hspace{0.05\linewidth}
    \subfloat[Unbalance Mutator\label{mnist_fedavg_unbalance}]{%
        \includegraphics[width=0.40\linewidth]{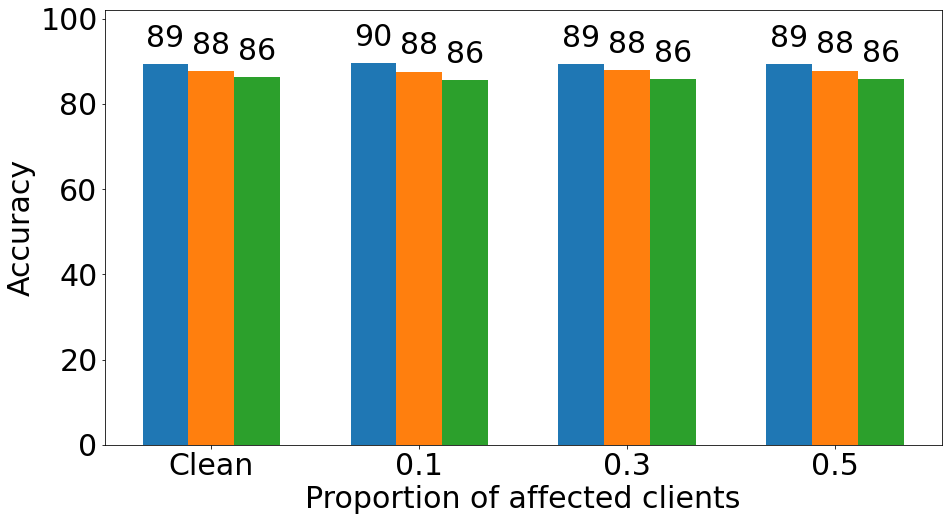}}
    \hfill
    \subfloat[Overlap Mutator\label{mnist_fedavg_overlap}]{%
        \includegraphics[width=0.40\linewidth]{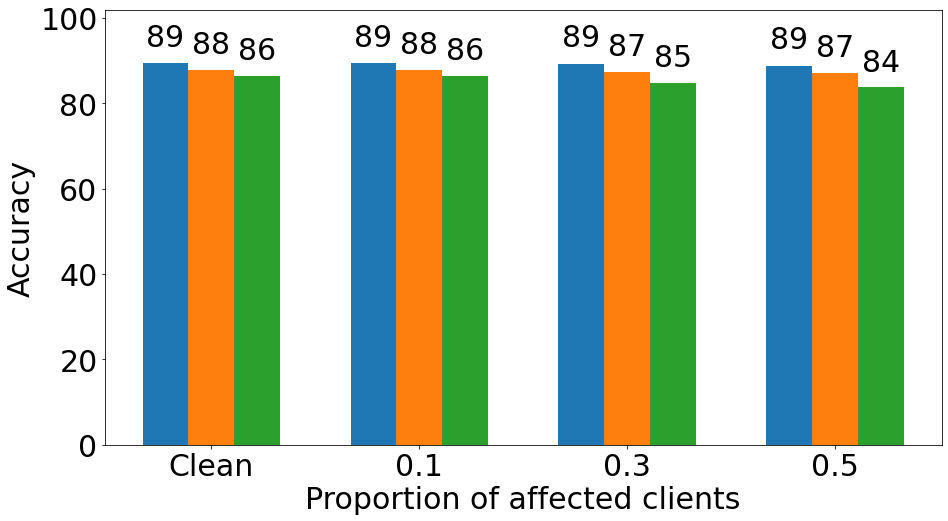}}
    \hspace{0.05\linewidth}
    \subfloat[Noise Mutator\label{mnist_fedavg_noise}]{%
        \includegraphics[width=0.40\linewidth]{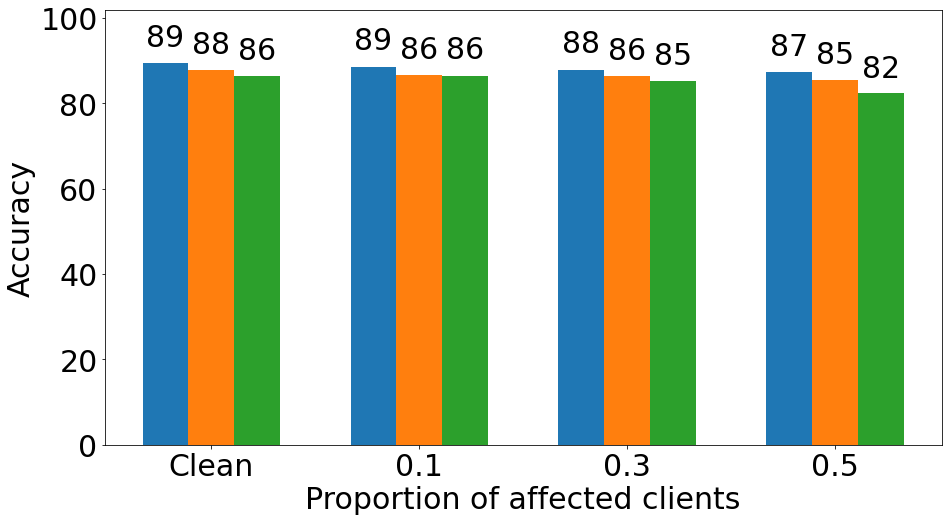}}
    \hfill
    \caption{Fashion MNIST - Federated Averaging performance under simulated faults.}
    \label{mnist_fedavg_faults_fig}
\end{figure*}

First, we discuss the results for faults simulated by mutation operators. 
The results for the CIFAR-10 and Fashion MNIST datasets are reported in Figures \ref{cifar_fedavg_faults_fig} and \ref{mnist_fedavg_faults_fig}, respectively.
Our first observation is that data mutators do not significantly impact the final accuracy of the model. Another interesting observation is that an increase in the proportion of affected clients does not noticeably decrease the accuracy. However, among these mutators, as Figures \ref{cifar_fedavg_noise} and \ref{mnist_fedavg_noise} show, the Noise mutator is more effective, and it gets slightly more powerful as the proportion increases. Therefore, a noisy dataset is a possible human fault that can cause problems in FL.

Moreover, a general and somewhat expected observation is that the accuracy slightly decreases as the dataset becomes more non-iid. This can be seen in different bar colors in Figures \ref{cifar_fedavg_faults_fig} and \ref{mnist_fedavg_faults_fig}.

\begin{table}

    \caption{The accuracy change (between clean and 0.5 proportion configs) of Federated Averaging, per mutator (averaged over all non-iid configurations).}
    \label{fedavg_faults_summary}
    \centering
    \begin{tabular}{cccccc}
    \bottomrule
        & \multicolumn{4}{c}{Accuracy change}\\
         & \multicolumn{2}{c}{CIFAR-10} & \multicolumn{2}{c}{Fashion MNIST}\\
        Mutator & Mean & Std & Mean & Std\\
        \toprule
        Delete & -4.3 & 0.7 &  -0.1 & 1.24\\
        Overlap & -0.71 & 3.29 & -1.3 & 4.11\\
        Unbalance & -0.64 & 1.01 & -0.21 & 1.75\\
        Noise & -5.66 & 1.05 & -2.84 & 1.04\\
    \bottomrule

    \end{tabular}
\end{table}

We report a summary of faults' effect on Federated Averaging in Table \ref{fedavg_faults_summary}. The change values reported are the amount of accuracy change between a clean scenario and the case where half of the clients are affected by the mutators. As the results show, the mutators do not significantly impact the accuracy. However, out of the mutators, the Noise mutator has the most impact on the final model, which is less than 6\% and is insignificant. Note that the statistical test results here show that in 20\% of the cases, these differences are statistically the same as well, but in 80\%, they are statistically different. However, the amount of difference itself is not actually significant (less than 6\%).

\textbf{Federated Averaging's performance under attacks:}

\begin{figure*}
    \centering
    \subfloat[CIFAR-10 - Label Flip\label{cifar_fedavg_label}]{%
        \includegraphics[width=0.40\linewidth]{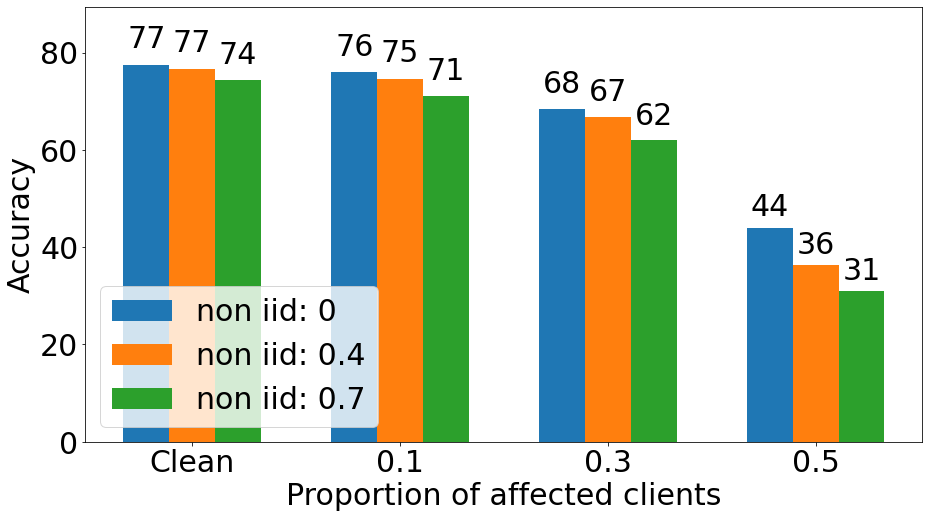}}
    \hspace{0.05\linewidth}
    \subfloat[Fashion MNIST - Label Flip\label{mnist_fedavg_label}]{%
        \includegraphics[width=0.40\linewidth]{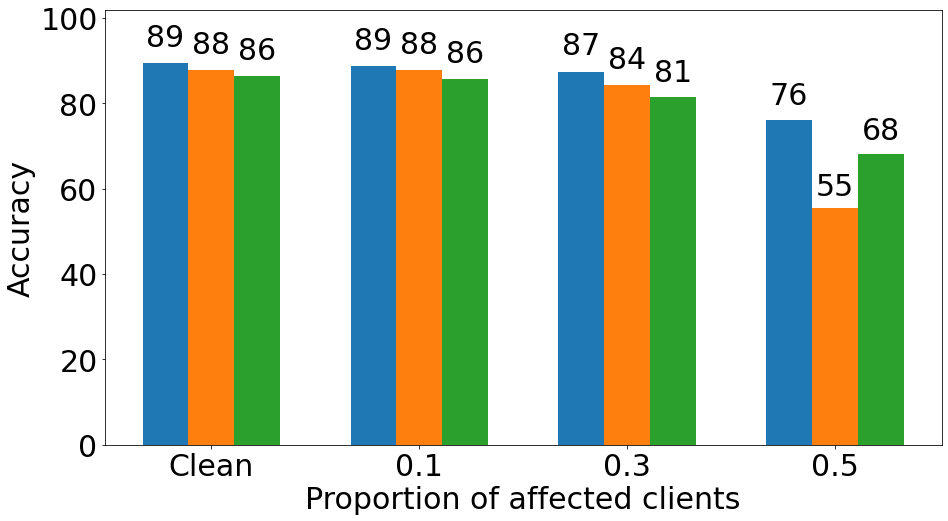}}
    \hfill
    \subfloat[CIFAR-10 - Random Update\label{cifar_fedavg_random}]{%
        \includegraphics[width=0.40\linewidth]{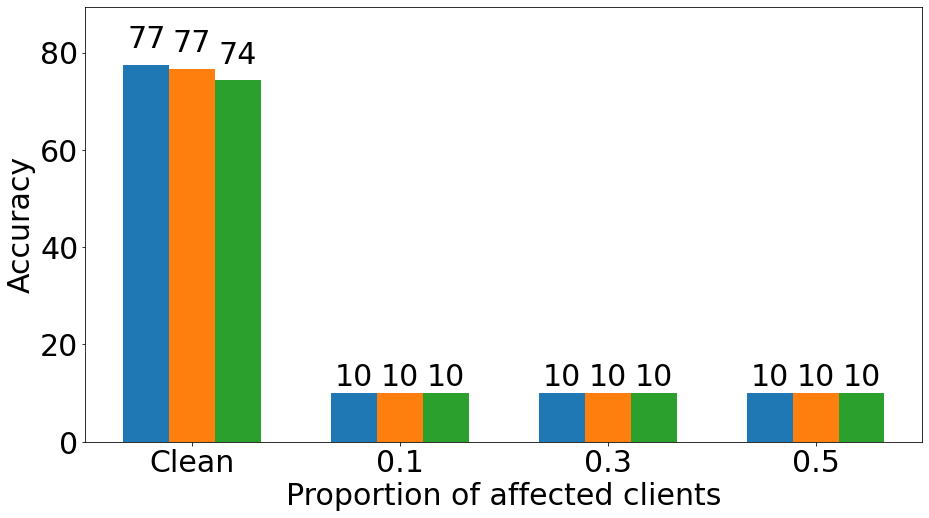}}
    \hspace{0.05\linewidth}    
    \subfloat[Fashion MNIST - Random Update\label{mnist_fedavg_random}]{%
        \includegraphics[width=0.40\linewidth]{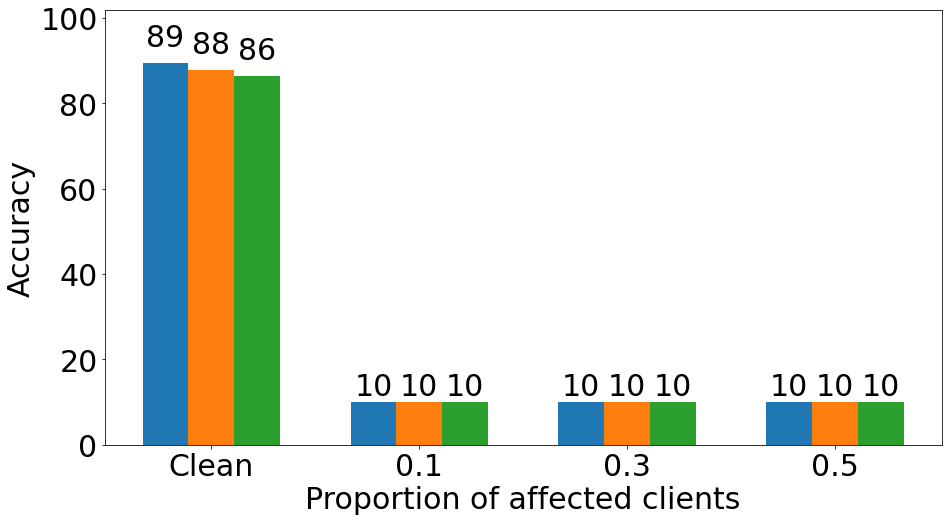}}
    \hfill
    \subfloat[CIFAR-10 - Sign Flip\label{cifar_fedavg_sign}]{%
        \includegraphics[width=0.40\linewidth]{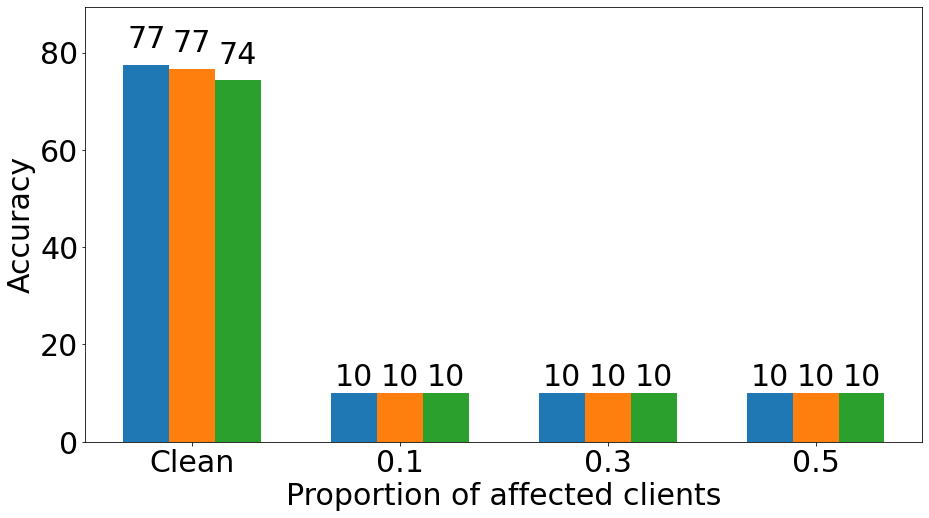}}
    \hspace{0.05\linewidth}    
    \subfloat[Fashion MNIST - Sign Flip\label{mnist_fedavg_sign}]{%
        \includegraphics[width=0.40\linewidth]{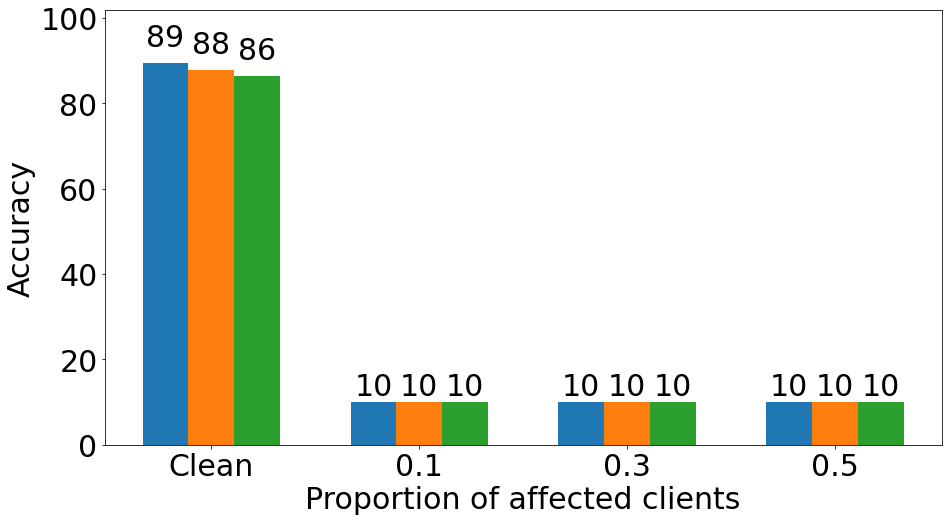}}
    \hfill
    \caption{Federated Averaging performance under untargeted attacks.}
    \label{fedavg_attacks_fig}
\end{figure*}

Figure \ref{fedavg_attacks_fig} shows the results of the untargeted attacks used in our experiments. As shown in Figures \ref{cifar_fedavg_label} and \ref{mnist_fedavg_label}, we can see that Label Flip is effective, and its effect is more noticeable in higher proportions like 0.3 and 0.5 (especially compared to the mutators discussed previously). Furthermore, a non-iid dataset can cause more damage as the proportion increases. On CIFAR-10, we see a 13\% decrease of accuracy (due to non-iid distribution) in 0.5 proportion compared to 5\% in 0.1 proportion. We see the same pattern on Fashion MNIST, and in 0.5 proportion, accuracy decreases by 8\%, but in 0.1 proportion, the decrease is around 3\%.  Considering that technically this is also a data mutator, human error and mislabelling samples can result in problems in FL.

Additionally, the results show that model poisoning attacks are much more powerful than data attacks and faults. As Figures \ref{cifar_fedavg_random} and \ref{cifar_fedavg_sign} show, even with a small proportion of affected clients, Federated Averaging loses its effectiveness completely, and the model classifies all test cases as a single class. As a result, accuracy reaches 10\%, which is similar to random guessing. The same can be seen from Figures \ref{mnist_fedavg_random} and \ref{mnist_fedavg_sign} for the Fashion MNIST dataset, and the model is essentially guessing randomly.

\begin{table}

    \caption{The accuracy change (between clean and 0.5 proportion configs) of Federated Averaging, per attack (averaged over all non-iid configurations).}
    \label{fedavg_attacks_summary}
    \centering
    \begin{tabular}{cccccc}
    \bottomrule
        & \multicolumn{4}{c}{Accuracy change}\\
         & \multicolumn{2}{c}{CIFAR-10} & \multicolumn{2}{c}{Fashion MNIST}\\
        Mutator & Mean & Std & Mean & Std\\
        \toprule
        Label Flip & -39.02 &7.81 & -21.26& 9.6\\
        Random Update & -66.13 & 1.67 & -77.81&1.53\\
        Sign Flip & -66.13 &1.67 & -77.81&1.53\\
    \bottomrule

    \end{tabular}
\end{table}

Like the faults section, we report a summary of attacks against Federated Averaging in Table \ref{fedavg_attacks_summary}. The change values reported are the amount of accuracy change between a clean scenario and the case where half of the clients are under attack.
Among the untargeted attacks, Random Update and Sign Flip are the most effective attacks against Federated Averaging, with around 70\% accuracy change. Furthermore, all the attacks are much more effective than mutators as they decrease the accuracy by at least 21.26\% (the best mutator did not even reach 6\%). Furthermore, the statistical test shows that all these changes are statistically different as well.

Regarding the Backdoor attack, we showed how it is applied in the background section in Figure \ref{backdoor_fig}.
In the CIFAR-10 dataset, images will be misclassified as a car with that specific pixel pattern. Also, in the Fashion MNIST dataset, they will be misclassified as a trouser.

\begin{figure*}
    \centering
    \subfloat[CIFAR-10\label{cifar_fedavg_backdoor}]{%
        \includegraphics[width=0.40\linewidth]{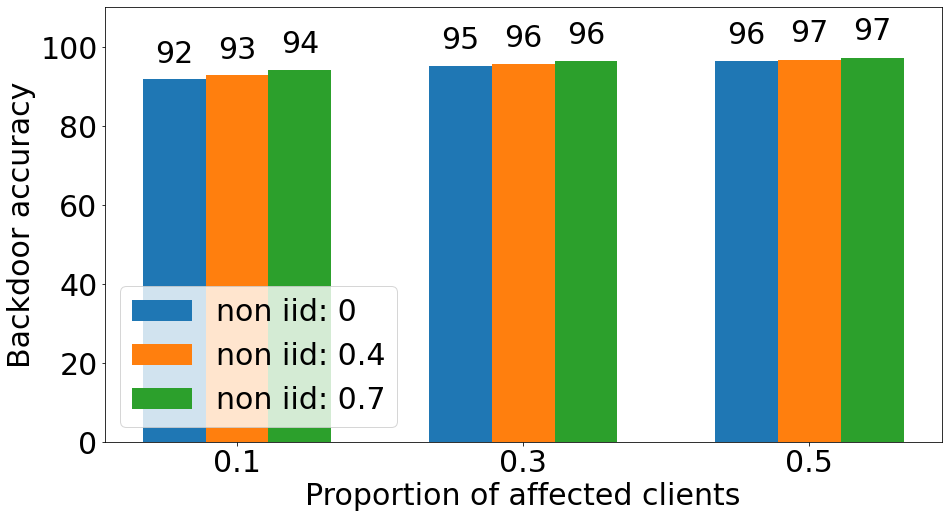}}
    \hspace{0.05\linewidth}
    \subfloat[Fashion MNIST\label{mnist_fedavg_backdoor}]{%
        \includegraphics[width=0.40\linewidth]{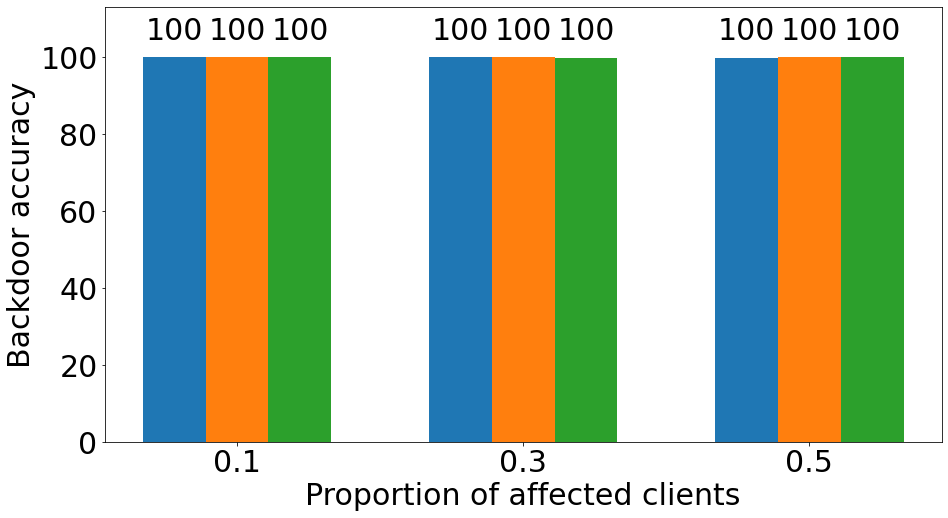}}
    \hfill
    \caption{Federated Averaging performance under the Backdoor attack.}
    \label{fedavg_backdoor_fig}
\end{figure*}

\begin{figure*}
    \centering
    \subfloat[Proportion=0.1\label{mnist_backdoor_0.1}]{%
        \includegraphics[width=0.40\linewidth]{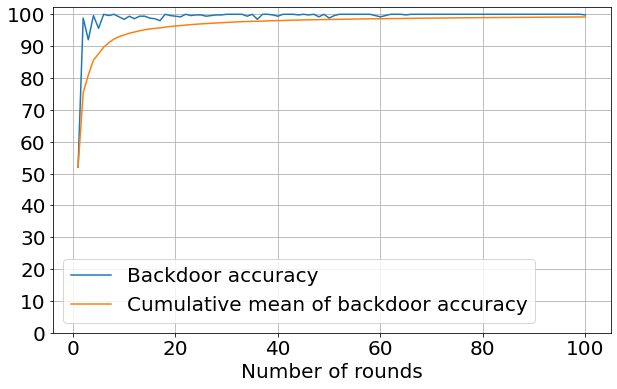}}
    \hspace{0.05\linewidth}
    \subfloat[Proportion=0.5\label{mnist_backdoor_0.5}]{%
        \includegraphics[width=0.40\linewidth]{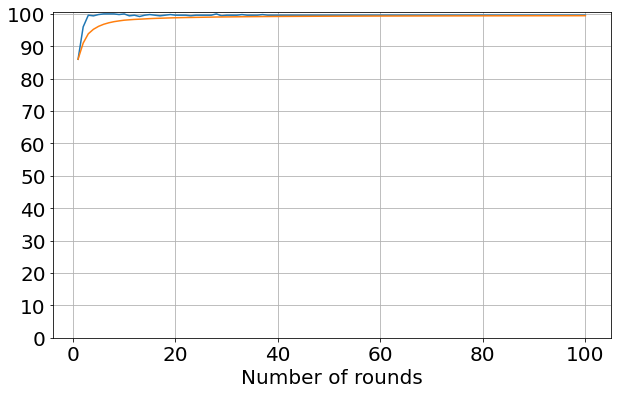}}
    \hfill
    \caption{Fashion MNIST - Backdoor task accuracy per round plots(iid distribution).}
    \label{mnist_backdoor_fig}
\end{figure*}

Finally, the results of the Backdoor attack are reported in Figure \ref{fedavg_backdoor_fig}. Note that the Backdoor results are reported separately from other attacks to avoid confusion. This is because other attack results report the accuracy of the main task, which the attack is trying to decrease. In contrast, Backdoor results report the backdoor task accuracy, which the attack is trying to increase.

Figure \ref{cifar_fedavg_backdoor} confirms that the Backdoor attack is highly effective on CIFAR-10.
As with even 0.1 proportion of malicious clients, the backdoor task reaches above 90\% accuracy in all non-iid cases.
This is more obvious in Figure \ref{mnist_fedavg_backdoor} where even with 0.1 proportion, the attacker reaches 100\% accuracy in all non-iid cases.
To clarify the Fashion MNIST case, we show the accuracy per round plots of Fashion MNIST in an iid scenario for 0.1 and 0.5 proportions in Figure \ref{mnist_backdoor_fig}. As we see, in a high proportion, the backdoor task accuracy starts from a very high value and converges to 100\% very fast, but in a small proportion, it starts from around 50\% and fluctuates a lot more before it converges. So the proportion has an effect here, but since the Fashion MNIST task is relatively simple, the backdoor task converges to maximum accuracy even in a small proportion in the end.

Also, in Figure \ref{cifar_fedavg_backdoor} for the CIFAR-10 dataset, we see that in the tiniest proportion, a more non-iid distribution increases the backdoor accuracy and makes it more powerful. This is not obvious in higher proportions as the backdoor reaches the highest feasible (considering the attack method and dataset) accuracy with an iid distribution. So, a more non-iid distribution cannot increase it. For the same reason, non-iid does not change the backdoor accuracy for the Fashion MNIST dataset as the backdoor accuracy is already 100\%.

\begin{tcolorbox}
\textbf{Answer to RQ1:} In generic image datasets (e.g., CIFAR-10 and Fashion MNIST), Federated Averaging is not robust against any attacks, and the FL process faces quality issues, as shown by the final model accuracy Also, generally, model attacks are more powerful than data attacks. Furthermore, data mutators do not significantly impact Federated Averaging, but noisy data leaves a bigger mark. Lastly, the non-iid distribution has a more detrimental impact on the quality of FL when more clients are attacked.
\end{tcolorbox}

\subsubsection{\textbf{RQ2 results (comparison of aggregators/defense mechanisms):}}
\label{results_rq2}

We divide this RQ into two parts: first, we compare the aggregators with no attack, then compare them under attacks.

\begin{figure}
    \centering
    \subfloat[CIFAR10\label{cifar_clean_fig}]{%
        \includegraphics[width=0.4\linewidth]{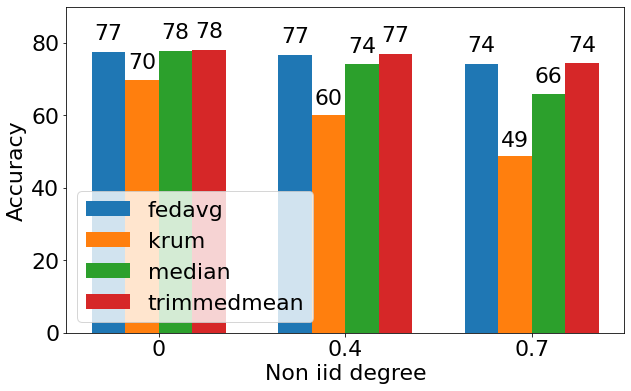}}
    \hspace{0.05\linewidth}
    \subfloat[Fashion MNIST\label{mnist_clean_fig}]{%
        \includegraphics[width=0.4\linewidth]{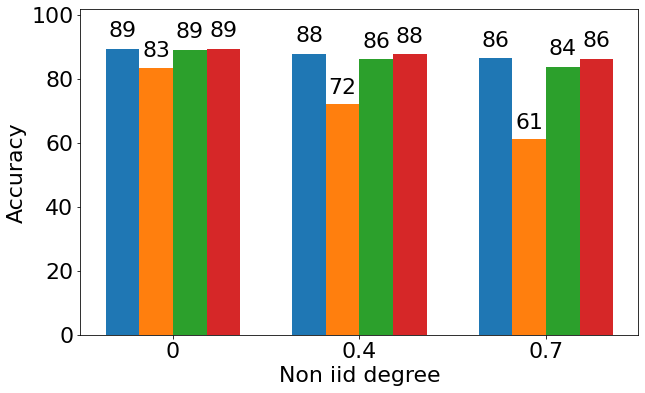}}
    \hfill
    \caption{Aggregators accuracy with no attack applied.}
    \label{clean_fig}
\end{figure}

We show the results for the first part in Figures \ref{cifar_clean_fig} and \ref{mnist_clean_fig} for CIFAR-10 and Fashion MNIST, respectively. Results show that Federated Averaging and Trimmed Mean perform similarly, and the non-iid does not significantly impact their accuracy. Median performs almost similarly to Federated Averaging and Trimmed Mean in a more iid setting, but it loses its accuracy slightly in a high non-iid situation.

In contrast, Krum's accuracy is noticeably lower than the other aggregators in an iid scenario. As the dataset becomes more non-iid, the difference between Krum and others becomes more significant.

Lastly, comparing two datasets shows that non-iid distribution decreases accuracy in CIFAR-10 more than Fashion MNIST, which can be because the former is more challenging than the latter.

\begin{figure*}
\captionsetup[subfloat]{farskip=2pt,captionskip=1pt}

    \centering
    \subfloat[Label Flip, non-iid=0\label{cifar_label_0}]{%
        \includegraphics[width=0.40\linewidth]{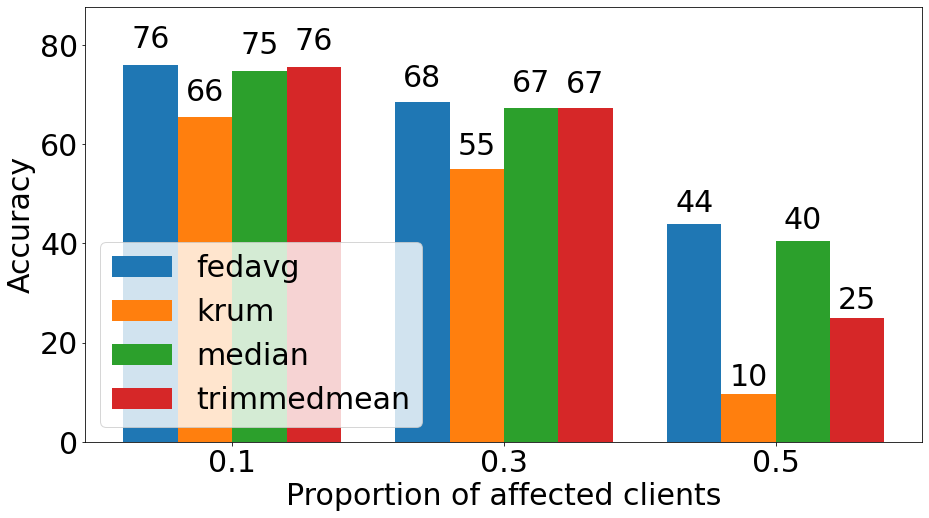}}
    \hspace{0.05\linewidth}
    \subfloat[Random Update, non-iid=0\label{cifar_random_0}]{%
        \includegraphics[width=0.40\linewidth]{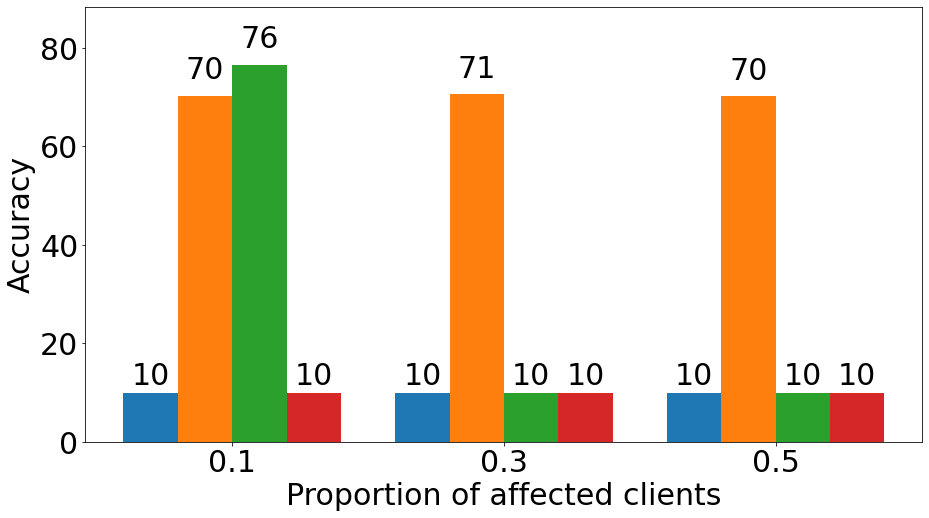}}
    \hfill
    \subfloat[Label Flip, non-iid=0.4\label{cifar_label_04}]{%
        \includegraphics[width=0.40\linewidth]{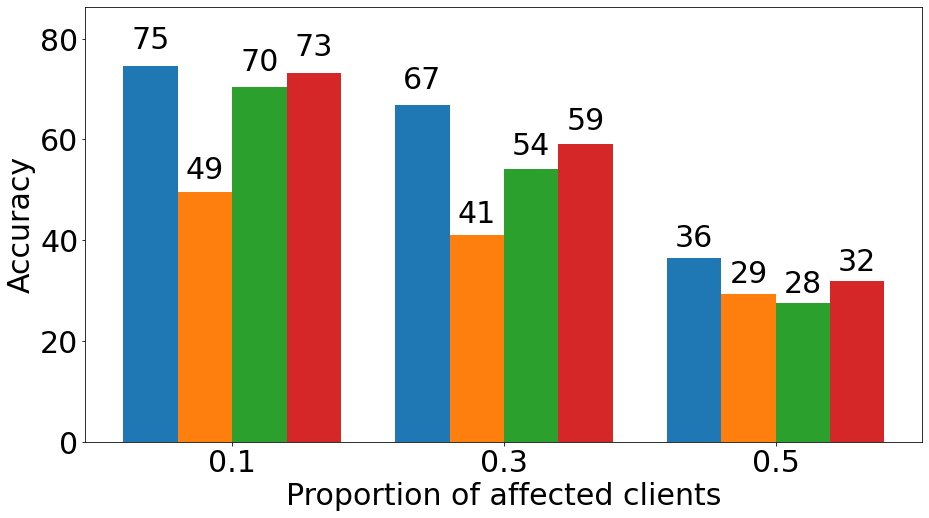}}
    \hspace{0.05\linewidth}
    \subfloat[Random Update, non-iid=0.4\label{cifar_random_04}]{%
        \includegraphics[width=0.40\linewidth]{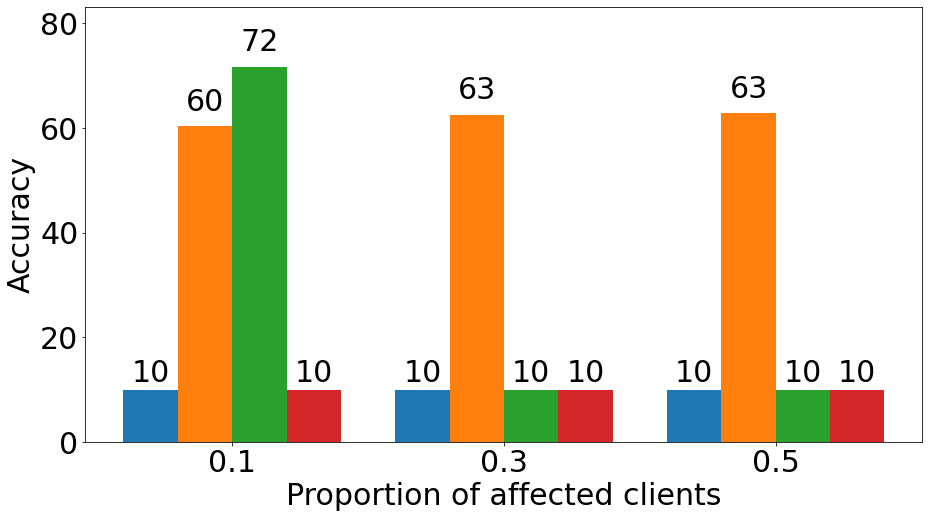}}
    \hfill
        \subfloat[Label Flip, non-iid=0.7\label{cifar_label_07}]{%
        \includegraphics[width=0.40\linewidth]{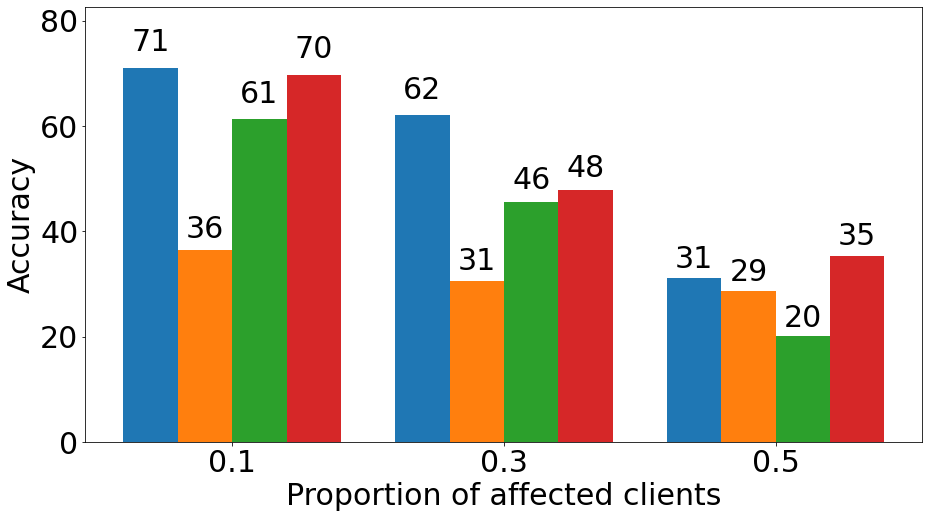}}
    \hspace{0.05\linewidth}
    \subfloat[Random Update, non-iid=0.7\label{cifar_random_07}]{%
        \includegraphics[width=0.40\linewidth]{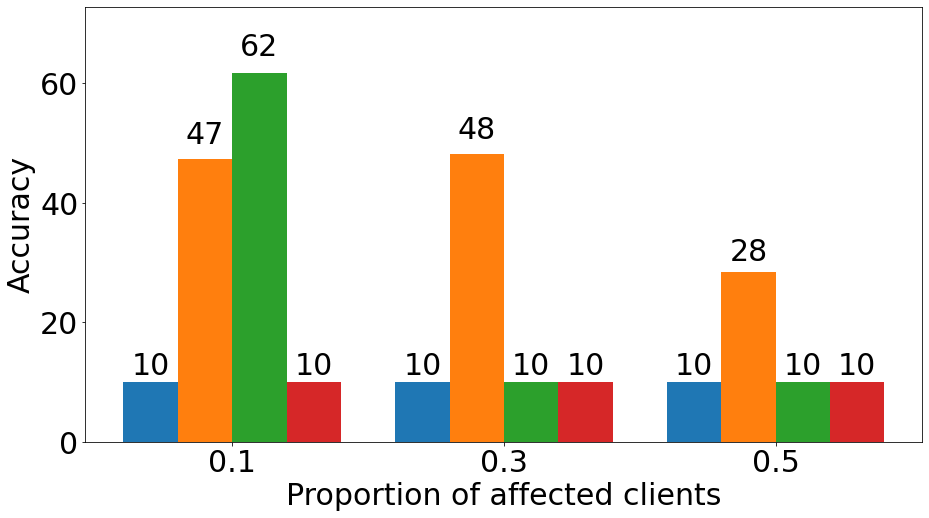}}
    \hfill
    \subfloat[Sign Flip, non-iid=0\label{cifar_sign_0}]{%
        \includegraphics[width=0.40\linewidth]{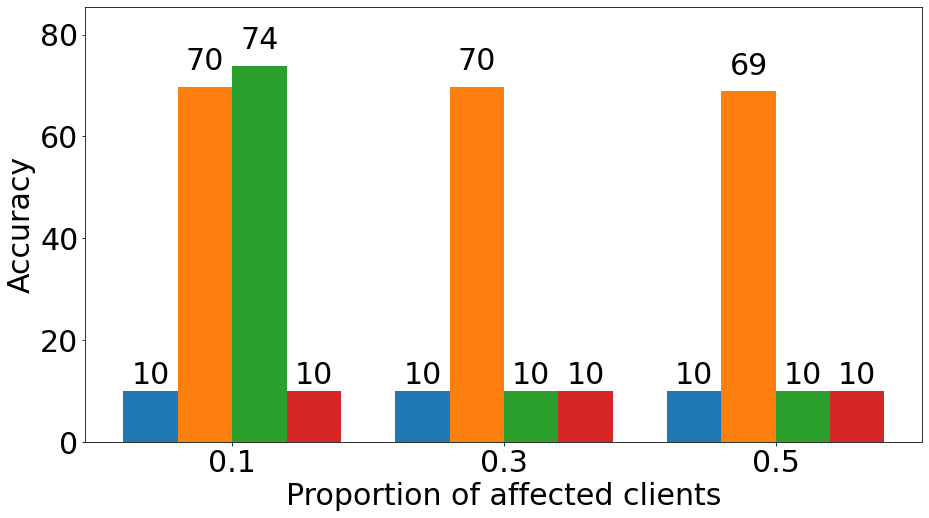}}
    \hspace{0.05\linewidth}
    \subfloat[Sign Flip, non-iid=0.4\label{cifar_sign_04}]{%
        \includegraphics[width=0.40\linewidth]{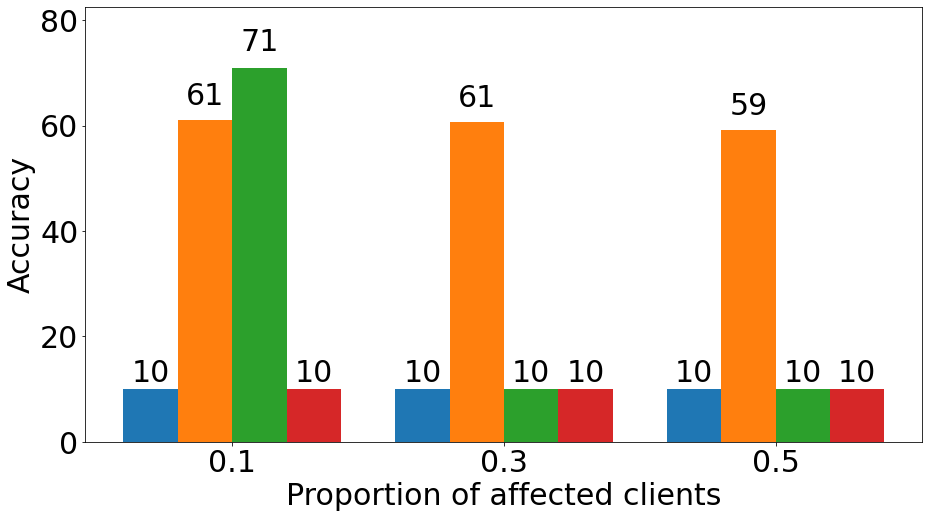}}
    \hfill
    \subfloat[Sign Flip, non-iid=0.7\label{cifar_sign_07}]{%
        \includegraphics[width=0.40\linewidth]{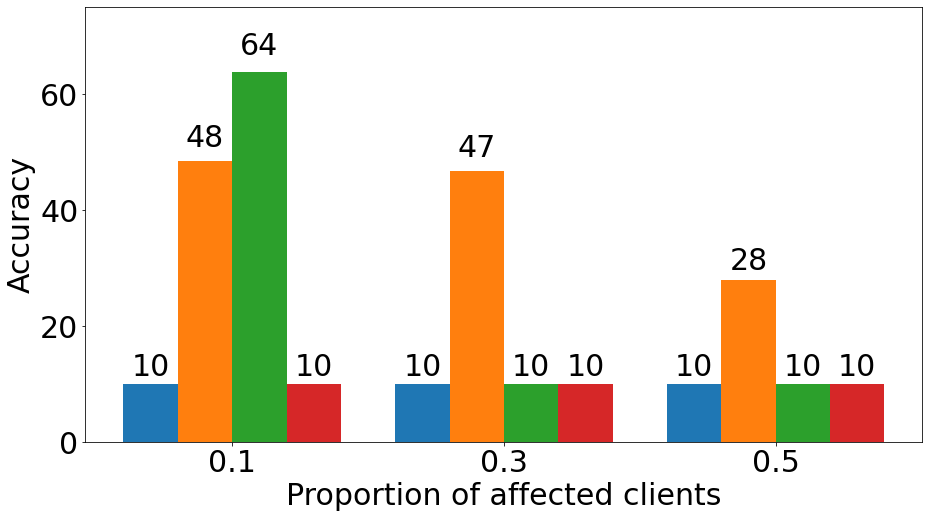}}
    \hfill    

    \caption{CIFAR-10 - Aggregators performance under untargeted attacks.}
    \label{cifar_aggregators_fig}
\end{figure*}

\begin{figure*}
\captionsetup[subfloat]{farskip=2pt,captionskip=1pt}

    \centering
    \subfloat[Label Flip, non-iid=0\label{mnist_label_0}]{%
        \includegraphics[width=0.40\linewidth]{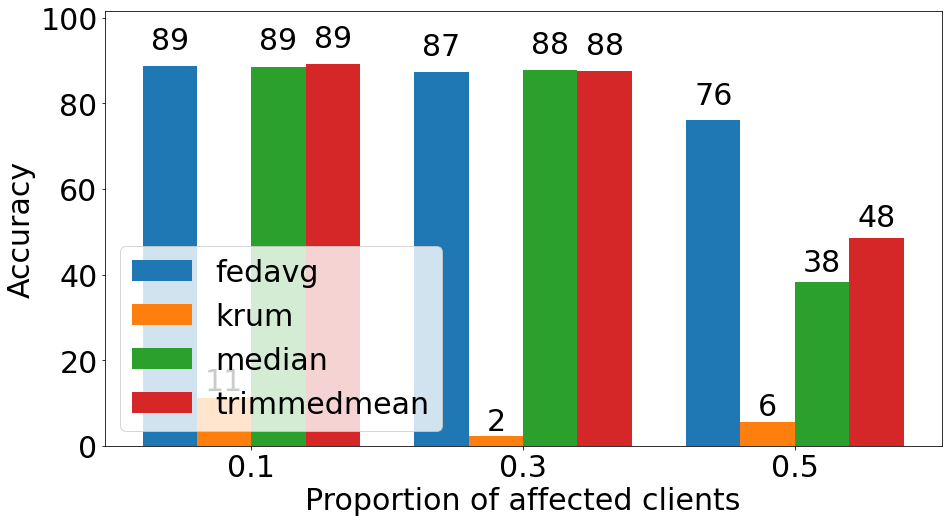}}
    \hspace{0.05\linewidth}
    \subfloat[Random Update, non-iid=0\label{mnist_random_0}]{%
        \includegraphics[width=0.40\linewidth]{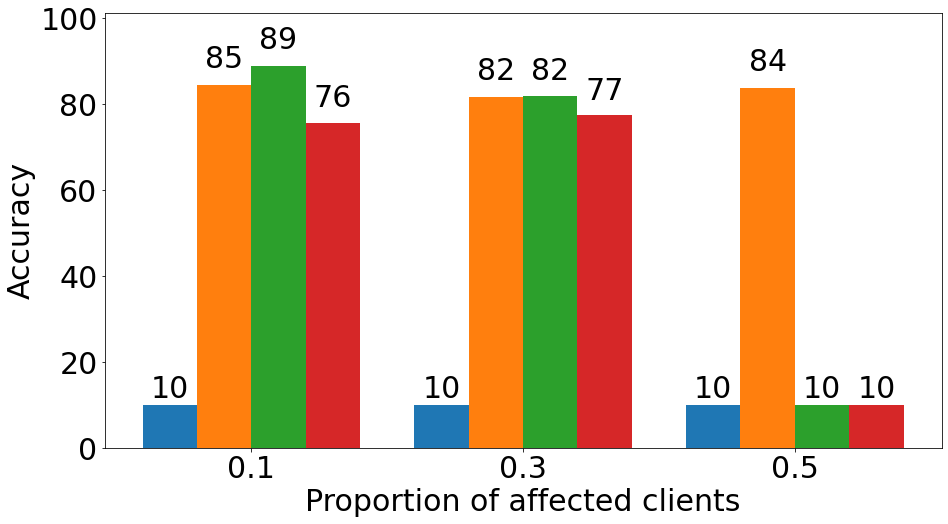}}
    \hfill
    \subfloat[Label Flip, non-iid=0.4\label{mnist_label_04}]{%
        \includegraphics[width=0.40\linewidth]{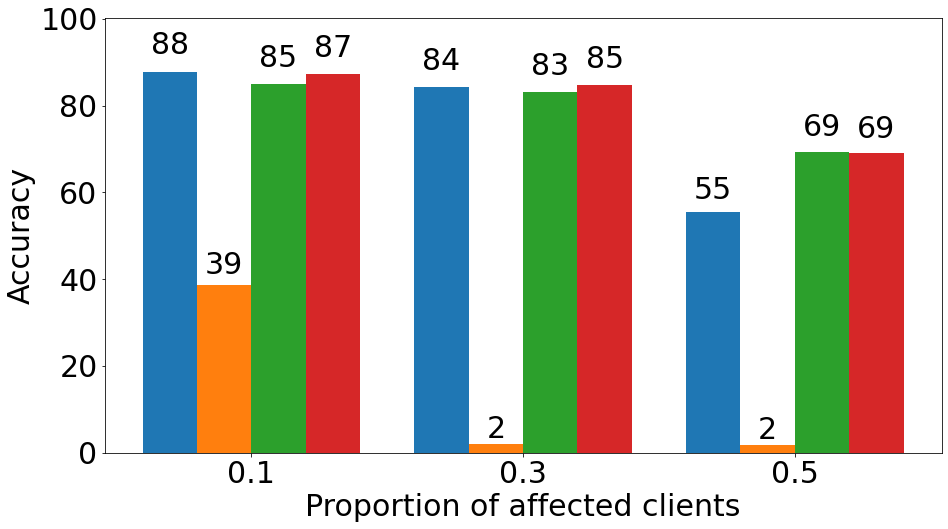}}
    \hspace{0.05\linewidth}
    \subfloat[Random Update, non-iid=0.4\label{mnist_random_04}]{%
        \includegraphics[width=0.40\linewidth]{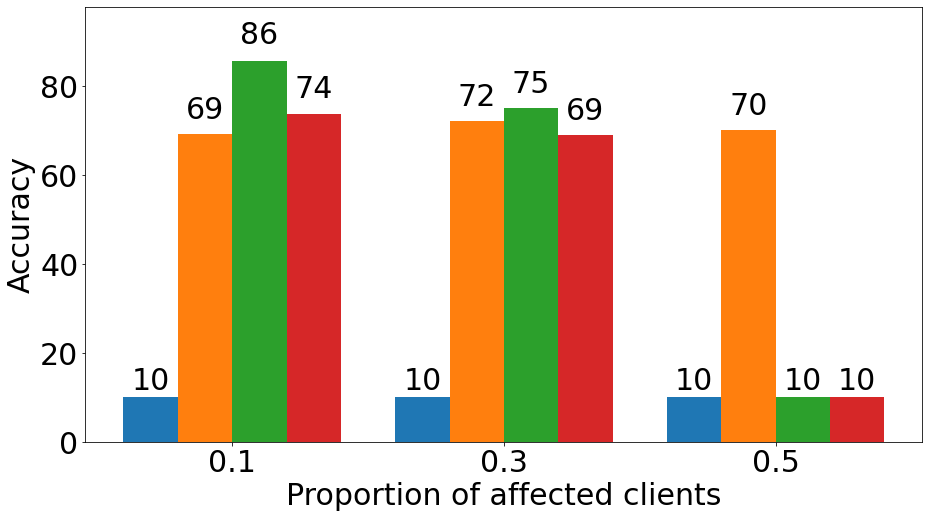}}
    \hfill
        \subfloat[Label Flip, non-iid=0.7\label{mnist_label_07}]{%
        \includegraphics[width=0.40\linewidth]{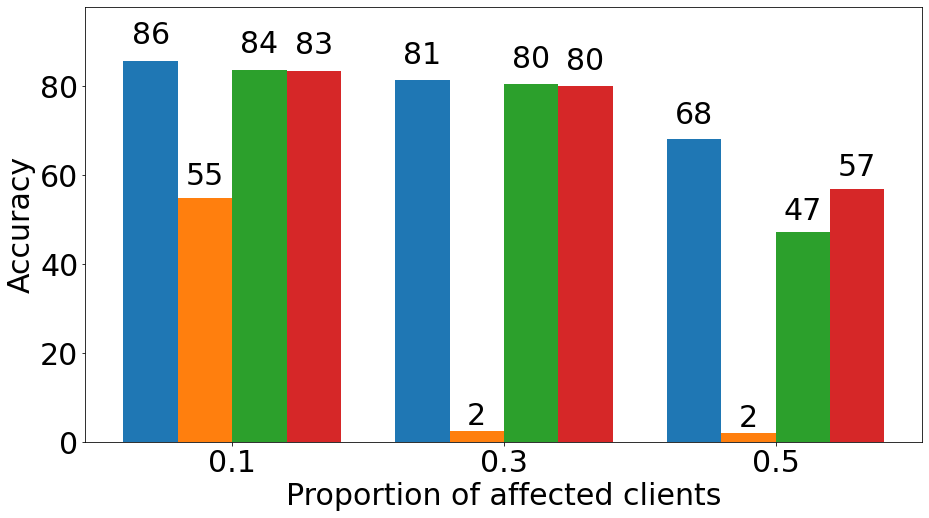}}
    \hspace{0.05\linewidth}
    \subfloat[Random Update, non-iid=0.7\label{mnist_random_07}]{%
        \includegraphics[width=0.40\linewidth]{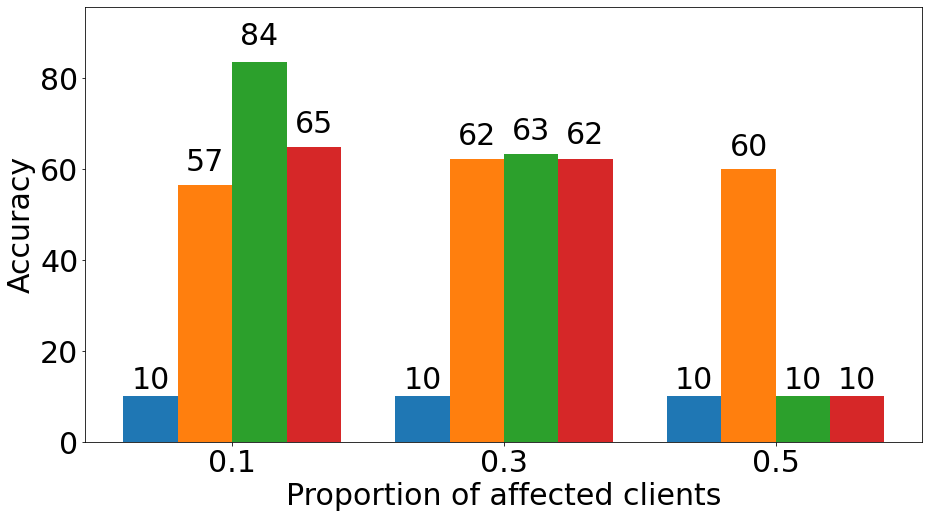}}
    \hfill
    \subfloat[Sign Flip, non-iid=0\label{mnist_sign_0}]{%
        \includegraphics[width=0.40\linewidth]{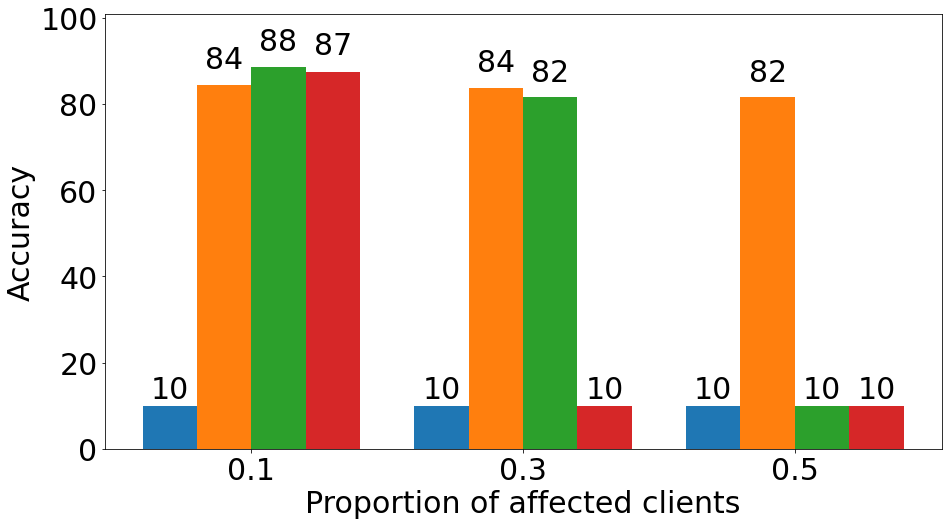}}
    \hspace{0.05\linewidth}
    \subfloat[Sign Flip, non-iid=0.4\label{mnist_sign_04}]{%
        \includegraphics[width=0.40\linewidth]{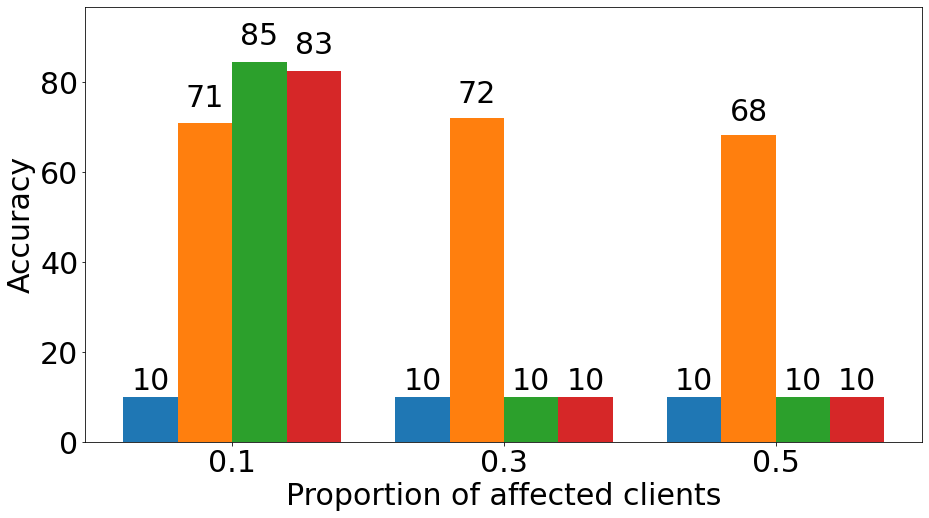}}
    \hfill
    \subfloat[Sign Flip, non-iid=0.7\label{mnist_sign_07}]{%
        \includegraphics[width=0.40\linewidth]{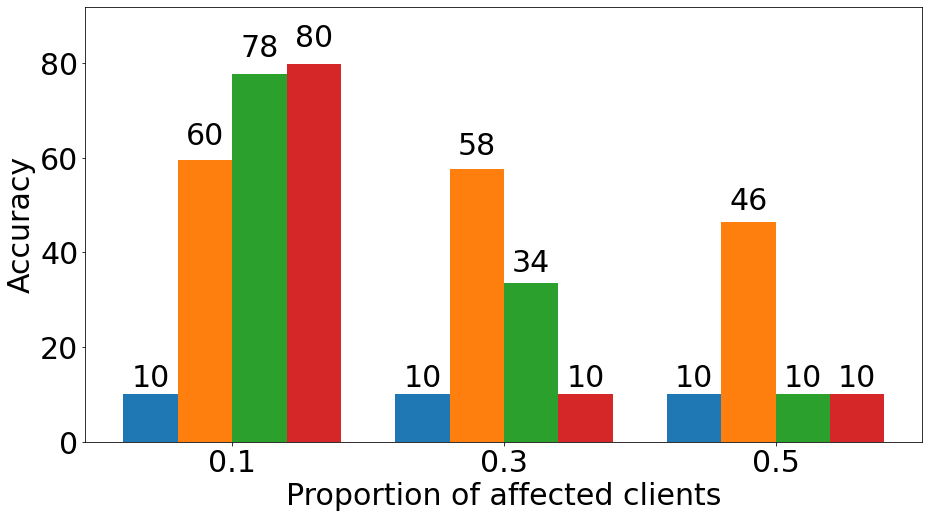}}
    \hfill    

    \caption{Fashion MNIST - Aggregators performance under untargeted attacks.}
    \label{mnist_aggregators_fig}
\end{figure*}

We compare the aggregators under various attacks for the second part of this research question. Since in RQ1, we saw that data mutators are not very effective against the baseline (Federated Averaging), we exclude them from this part.
Figures \ref{cifar_aggregators_fig} and \ref{mnist_aggregators_fig} show the results of untargeted attacks for the CIFAR-10 and Fashion MNIST, respectively. For the reasons discussed in RQ1, separate Backdoor attack results from untargeted attack results.

As Figure \ref{cifar_label_0} shows, in the Label Flip attack on CIFAR-10, when we consider an iid situation, Federated Averaging, Median, and Trimmed Mean perform very similarly in smaller attack proportions. However, Trimmed Mean falls behind when half of the clients are byzantine. However, Krum does not perform well in the Label Flip attack and roughly achieves 10\% lower accuracy in smaller proportions than other methods; this gets much worse when half of the clients are attacked.

Figure \ref{mnist_label_0} shows similar patterns on the Fashion MNIST dataset in an iid case. We see that Federated Averaging, Median, and Trimmed Mean are very similar in small proportions, but in 0.5 proportion, Federated Averaging is noticeably more robust. Like CIFAR-10, Krum is the worst aggregator by far, and its results are even worse than CIFAR-10 as it reaches as low as 2\% accuracy in some cases.

Looking at Figures \ref{cifar_label_04} and \ref{cifar_label_07} for the CIFAR-10 dataset, as the non-iid degree increases, Federated Averaging outperforms other techniques in almost all the cases, and its advantage is more noticeable, with 30\% of malicious clients.
We also see that Trimmed Mean gets better results than the Median in almost all cases, making it a better choice than the Median in a non-iid scenario.
Krum gets even worse when data becomes more non-iid, and even with 0.1 malicious clients, it works way worse than when it was not attacked.
Consequently, although Federated Averaging is not robust against this attack, it still is the best choice and generally works better than the others.

On the Fashion MNIST dataset, Figures \ref{mnist_label_04} and \ref{mnist_label_07} show similar findings, and Federated Averaging and Trimmed Mean are the best choices.

To summarize the Label Flip attack, generally, robust aggregators perform worse than Federated Averaging in a non-iid case when they are under attack. This might be because the robust aggregators cannot distinguish what is causing the difference between the updates (high non-iid or the attack itself), which results in poor performance.

Looking at the untargeted model poisoning attacks (Random Update and Sign Flip) for the CIFAR-10 dataset. The first remark is that even with the smallest proportion of affected clients in an iid case (Figures \ref{cifar_random_0} and \ref{cifar_sign_0}), Federated Averaging and Trimmed Mean are entirely ineffective, and the model is guessing randomly. Moreover, when a small proportion of clients are affected, Median works best in all non-iid degrees and achieves the highest accuracy (Figures \ref{cifar_random_04}, \ref{cifar_random_07} and \ref{cifar_sign_04}, \ref{cifar_sign_07}). Nevertheless, when more clients become byzantine, it too loses its effectiveness and surrenders to attackers like Federated Averaging and Trimmed Mean. In contrast to the label attack where Krum struggled, here is where Krum shines. In non-iid degrees of 0 and 0.4, Krum shows incredible robustness against the attacks. No matter the proportion of affected clients, it performs excellently and achieves results as if it were not under attack. However, in the highest non-iid case, Krum relatively loses its robustness when half of the clients become byzantine.

The observations are a bit different on the Fashion MNSIT dataset. Firstly, we see that Trimmed Mean is still viable against Random Update attack in all non-iid cases under small proportions (Figures \ref{mnist_random_0}, \ref{mnist_random_04}, \ref{mnist_random_07}). This is because the Fashion MNIST task is less complex compared to CIFAR-10. However, in the Sign Flip attack, Trimmed Mean is guessing randomly in more cases compared to the Random Update attack (Figures \ref{mnist_sign_0}, \ref{mnist_sign_04}, \ref{mnist_sign_07}), which confirms the fact that the Sign Flip attack is more powerful. Everything else is similar to CIFAR-10, and Median and Krum are still the best options.

To sum up, in untargeted attacks, generally speaking, Krum is the most robust technique and works very reliably in all cases. However, if the proportion of affected clients is small, the Median is better because it has better convergence speed and accuracy than Krum.

\begin{figure*}
\captionsetup[subfloat]{farskip=2pt,captionskip=1pt}

    \centering
    \subfloat[CIFAR-10, non-iid=0\label{cifar_backdoor_0}]{%
        \includegraphics[width=0.40\linewidth]{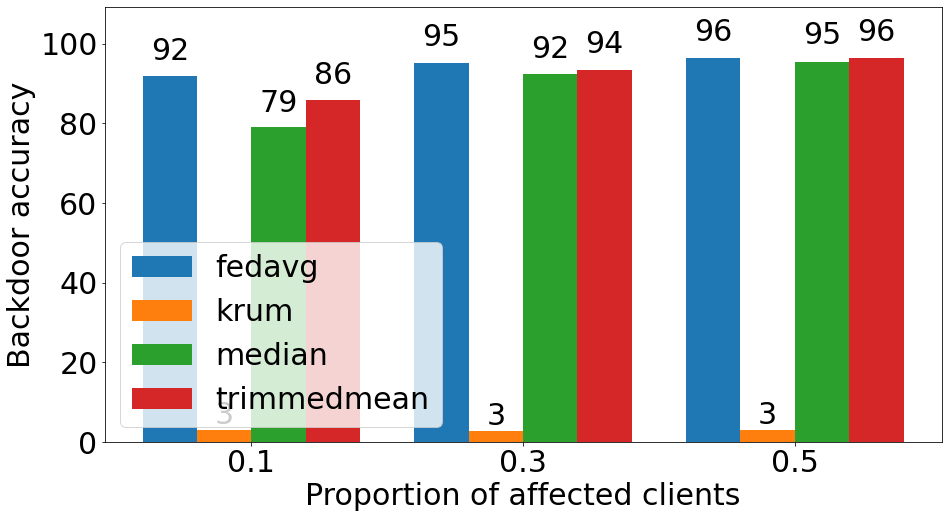}}
    \hspace{0.05\linewidth}
    \subfloat[Fashion MNIST, non-iid=0\label{mnist_backdoor_0}]{%
        \includegraphics[width=0.40\linewidth]{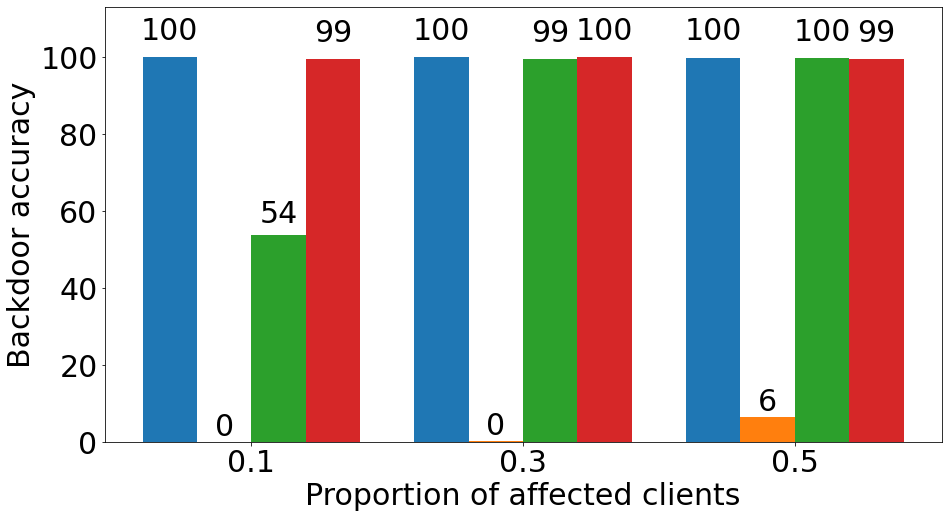}}
    \hfill
    \subfloat[CIFAR-10, non-iid=0.4\label{cifar_backdoor_04}]{%
        \includegraphics[width=0.40\linewidth]{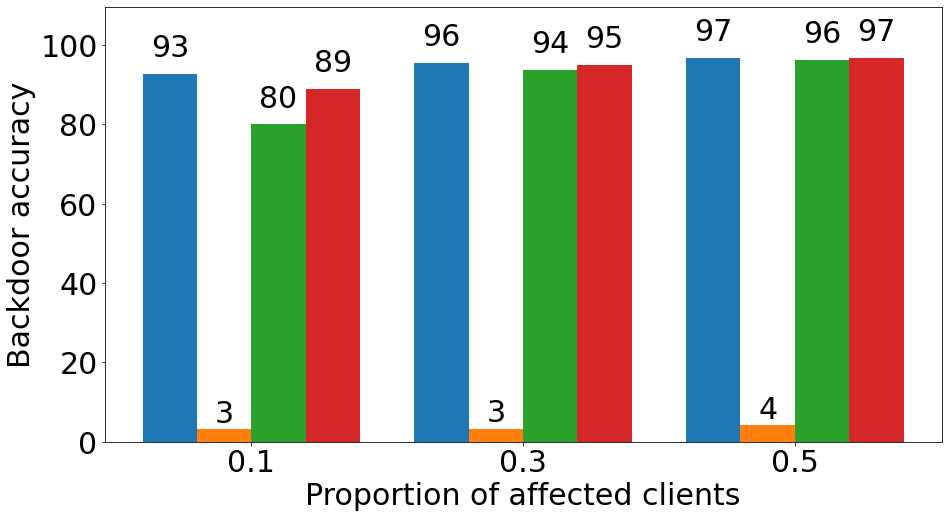}}
    \hspace{0.05\linewidth}
    \subfloat[Fashion MNIST, non-iid=0.4\label{mnist_backdoor_04}]{%
        \includegraphics[width=0.40\linewidth]{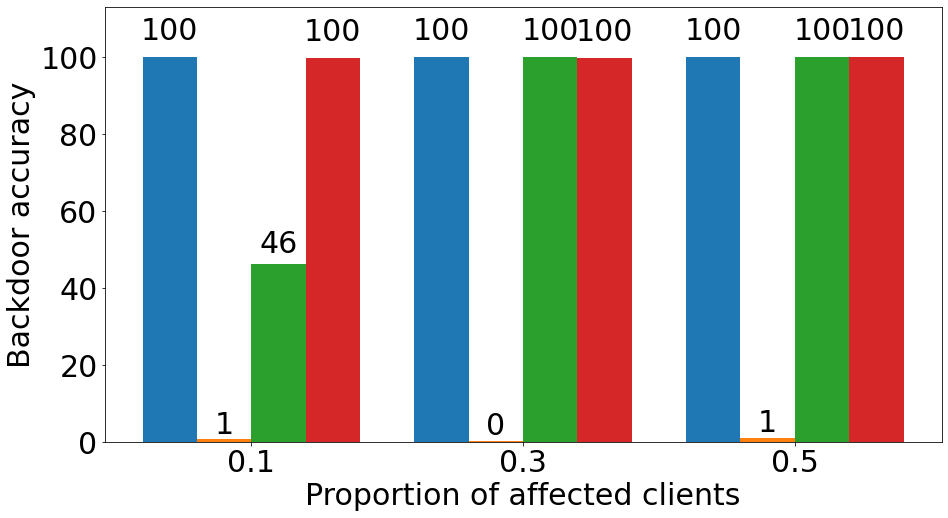}}
    \hfill
    \subfloat[CIFAR-10, non-iid=0.7\label{cifar_backdoor_07}]{%
        \includegraphics[width=0.40\linewidth]{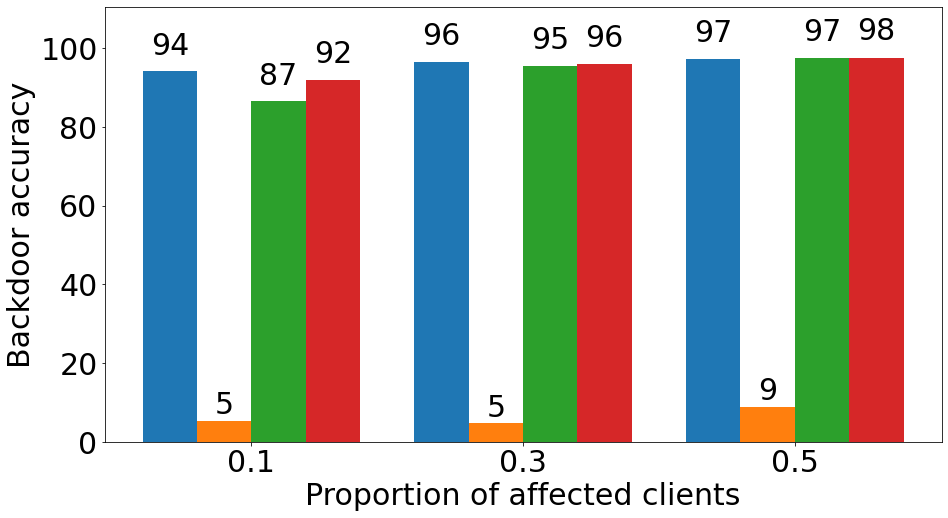}}
    \hspace{0.05\linewidth}
    \subfloat[Fashion MNIST, non-iid=0.7\label{mnist_backdoor_07}]{%
        \includegraphics[width=0.40\linewidth]{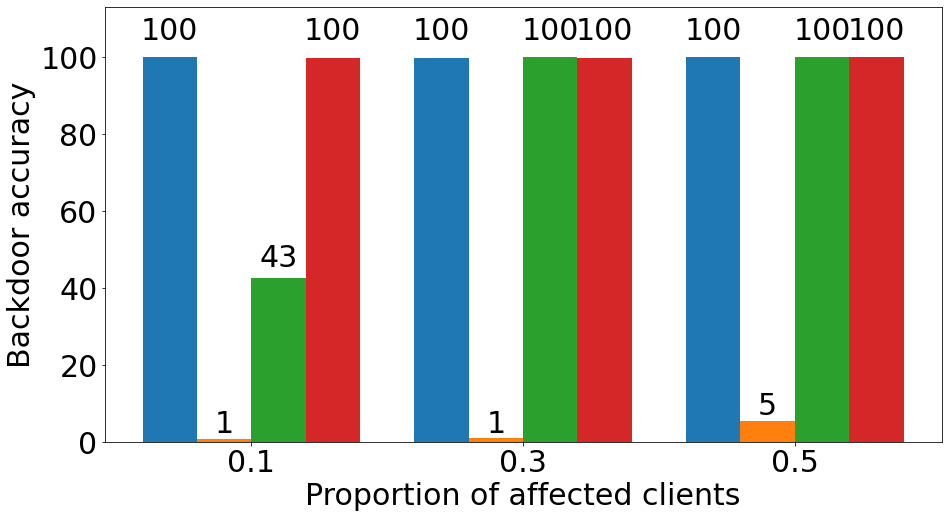}}
    \hfill    

    \caption{Aggregators performance under the Backdoor attack.}
    \label{backdoor_aggregators_fig}
\end{figure*}

The results for the Backdoor attack are shown in Figure \ref{backdoor_aggregators_fig}. As it can be seen, Krum is quite robust in all cases. Moreover, with the increase of attackers or the non-iid degree, the final model does not get fooled, and backdoor accuracy is close to zero. On the other hand, Federated Averaging gets entirely fooled in all cases and is not robust. Median and Trimmed Mean show some resilience against the Backdoor attack. The former is slightly superior when data is more iid, and the proportion of affected clients is small. However, they are not even close to Krum, and as the proportion increases, they become ineffective like Federated Averaging, and the backdoor goal is almost always achieved.
In this case, Krum is undoubtedly the best and most robust technique.

\begin{table}

    \caption{Aggregators summary in terms of their accuracy (averaged across attacks, non-iid configurations, and proportion of affected clients) and the number of times the aggregator is the best choice among all aggregators under study (Number of times achieving the Top rank) -- The shaded cells mark the best techniques.}
    \label{cifar_summary}
    \centering
    \subfloat[Results for untargeted attacks (higher accuracy is better)]{
\begin{tabular}{ccccc}
    \bottomrule
         &  & \multicolumn{2}{c}{Accuracy} & \multirow{2}{*}{\shortstack{Top rank\\frequency}}\\
        Dataset  & Aggregator & Mean&Std &\\ 
        \toprule
        \multirow{4}{*}{CIFAR-10}    &   FedAvg & 26.31 &24.88& 8\\
&    Krum                & \cellcolor{gray!25}51.09 & \cellcolor{gray!25}16.66 & \cellcolor{gray!25}12\\
&    Median              & 37.04 & 27.32 & 6\\
&    Tri-mean        & 24.61 & 23.2 & 1\\

\midrule

\multirow{4}{*}{\shortstack{Fashion \\ MNIST}}   &   FedAvg & 33.14 & 33.29 & 5\\
&   Krum                & 51.32 & \cellcolor{gray!25}30.27 & 9\\
&    Median              & \cellcolor{gray!25}58.43 & 31.93 & \cellcolor{gray!25}10\\
&  Tri-mean        & 53.69  & 32.25 & 3\\

    \bottomrule
    \end{tabular}
    }
\vspace{0.07\linewidth}
\subfloat[Results for the Backdoor attack (lower backdoor accuracy is better)]{
\begin{tabular}{ccccc}
    \bottomrule
         &  & \multicolumn{2}{c}{Backdoor accuracy} & \multirow{2}{*}{\shortstack{Top rank\\frequency}}\\
        Dataset  & Aggregator & Mean&Std &\\ 
        \toprule
        \multirow{4}{*}{CIFAR-10}    &   FedAvg & 95.16 & \cellcolor{gray!25}1.78 & 0\\
&    Krum                & \cellcolor{gray!25}4.26 & 1.86 & \cellcolor{gray!25}9\\
&    Median              & 90.72 & 6.65& 0\\
&    Tri-mean        & 93.56 & 3.71& 0\\

\midrule

\multirow{4}{*}{\shortstack{Fashion \\ MNIST}}   &   FedAvg & 99.94 & \cellcolor{gray!25}0.08 & 0\\
&    Krum                & \cellcolor{gray!25}1.71 & 2.27 & \cellcolor{gray!25}9\\
&   Median              & 82.39 & 24.79 & 0\\
&   Tri-mean        & 99.73 & 0.23& 0\\

    \bottomrule
    \end{tabular}
    }

\end{table}

We report a summary of aggregators for CIFAR-10 and Fashion MNIST in Table \ref{cifar_summary} based on the average accuracy of the final model on the test data and the number of times each aggregator was the most robust one. Note that only attacks are selected here since data mutators were ineffective and untargeted attacks and backdoor attacks were split to avoid confusion. As it can be seen, Krum achieves the best accuracy on average and achieves the top rank the most in the CIFAR-10 dataset, so all in all, it is the most robust aggregation method for that dataset. For Fashion MNIST, Median is the most robust aggregator. However, considering the first rank count of aggregators, Krum comes in a close second, but its mean accuracy is less than of the Median. This is because it achieves far worse results in the Label Flip attack.

Lastly, we report the statistical tests. We have 2 datasets, 4 attacks, 3 proportions, and 3 non-iid degrees, which results in 72 cases. Based on the previous discussions, we consider Krum to be the best option overall and compare it to the second-best aggregator (per case) in all the cases. 
The results show that in 38 cases, Krum is the best aggregator with a statistically significant difference. In one case, it is the best, but the difference is insignificant. It is not the best in two cases, but the difference is not significant either. Finally, in 31 cases, Krum is worse with a significant difference. As a result, we can say that Krum is the best choice in 57\% of the cases.
Doing the same test for Median, Trimmed Mean, and Federated Averaging will result in 52\%, 45.8\%, and 50\%. These results confirm that Krum is the best aggregator overall.

\begin{tcolorbox}
\textbf{Answer to RQ2:}
In 57\% of the cases, Krum is the most robust FL aggregator, and Median comes in second when there is no prior knowledge of the type of attack or fault (which is often the case). However, if attacks were known, the best choice for the Label Flip attack would still be Federated Averaging. Also, if the model was under untargeted model attacks and the number of attackers was small, Median would be a better choice than Krum. That being said, Krum's main drawback is its problem with non-iid distribution, which struggles to maintain the same accuracy as the iid degree increases. So, in short, there is no best defense technique for all situations to ensure quality.
\end{tcolorbox}

\subsubsection{\textbf{RQ3 results (the case of a real federated dataset):}
}
\label{results_rq3}

\begin{figure}
    \centering
    \includegraphics[width=0.4\textwidth]{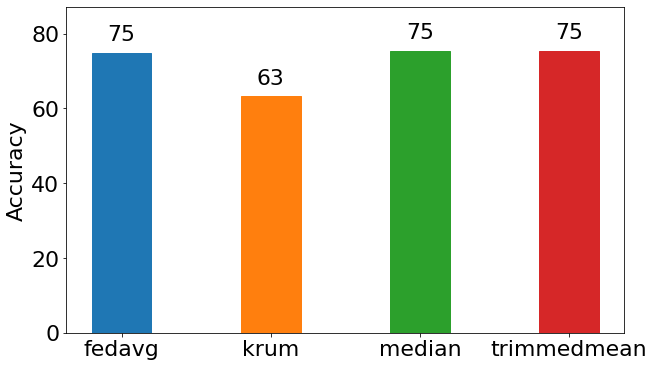}
    \caption{ADNI - Aggregators accuracy with no attack.}
    \label{adni_clean_fig}
\end{figure}

Like RQ2, we first report the results where all clients are benign in Figure \ref{adni_clean_fig}. As it can be seen, all aggregators perform similarly and achieve the same accuracy of 75\% except Krum. This is much like the results we saw for general datasets in RQ2. However, there is a significant difference in the ADNI dataset for Krum. Here Krum does not perform reliably and shows different behavior with different runs.
Moreover, Krum converges to very different results, and there is a 15\% difference in the final accuracy for Krum's worst and best run. Whereas in CIFAR-10, the variation between the results is less than 5\%. This inconsistency might be because this dataset is naturally non-iid and unbalanced.

\begin{figure}
    \centering
    \subfloat[ADNI - Worst\label{krum_worst_adni}]{
        \includegraphics[width=0.4\linewidth]{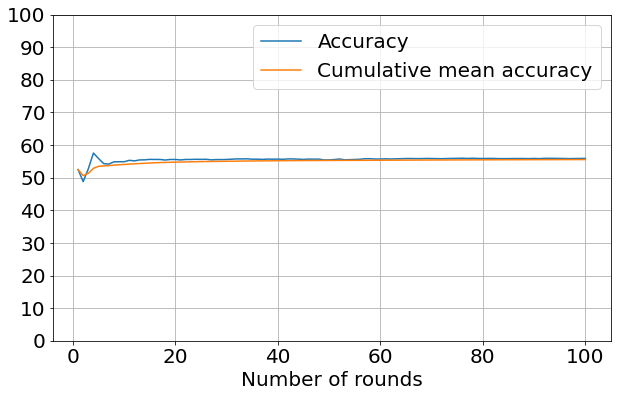}}
    \hspace{0.05\linewidth}
    \subfloat[ADNI - Best\label{krum_best_adni}]{
        \includegraphics[width=0.4\linewidth]{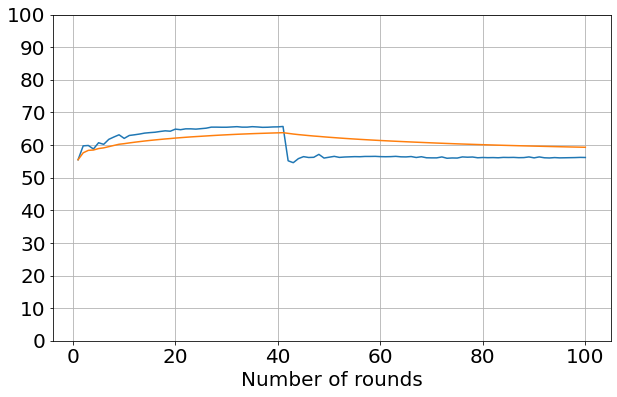}}
    \hfill
    \subfloat[CIFAR-10 - Worst (non-iid=0.4)\label{krum_worst_cifar}]{
        \includegraphics[width=0.4\linewidth]{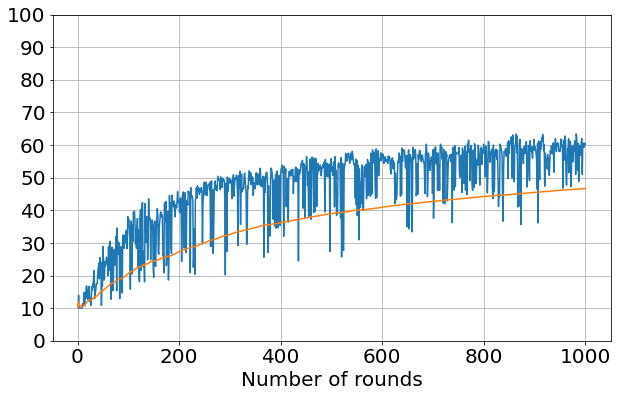}}
    \hspace{0.05\linewidth}
    \subfloat[CIFAR-10 - Best (non-iid=0.4)\label{krum_best_cifar}]{
        \includegraphics[width=0.4\linewidth]{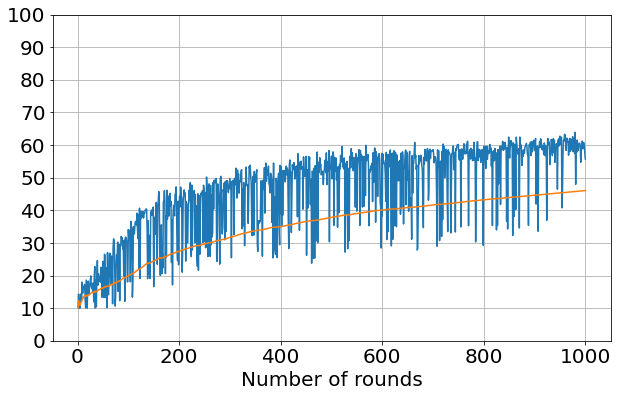}}
    \hfill
    \caption{Convergence plots of Krum on different runs.}
    \label{adni_krum_fig}
\end{figure}

We saw in RQ2 that Krum performed worse compared to others in synthetic non-iid cases, and here we see a similar problem but more severe.
To illustrate this better, Figure \ref{adni_krum_fig} shows Krum's test accuracy convergence trend over the communication rounds for ADNI and CIFAR-10 datasets on worst and best runs. As seen in Figures \ref{krum_worst_adni} and \ref{krum_best_adni}, Krum does not work reliably here, and it converges to very different results, and we see a 15\% difference in the final accuracy. Whereas in CIFAR-10, the results difference is less than 5\%.
One explanation for this inconsistency is that this RQ is investigating a cross-silo FL case, meaning clients are not selected randomly. Thus, Krum always sees updates from the same clients and chooses one which might be good or bad for the final test.
Since different centers have images from different MRI machines, the selected client update by Krum might not be a generally good solution, and final results are dependent on random factors like the data shuffle method, weight initialization, and other factors. Consequently, Krum is not guaranteed to converge to the best solution in this dataset, and its results are unreliable and quite random.

Note that the high fluctuations visible in Figures \ref{krum_best_cifar} and \ref{krum_worst_cifar} are because Krum only selects one client's update at each round, and the setting shown here is non-iid and cross-device, resulting in the fluctuations.

\begin{figure*}
    \centering
    \subfloat[Delete Mutator\label{adni_delete}]{%
        \includegraphics[width=0.40\linewidth]{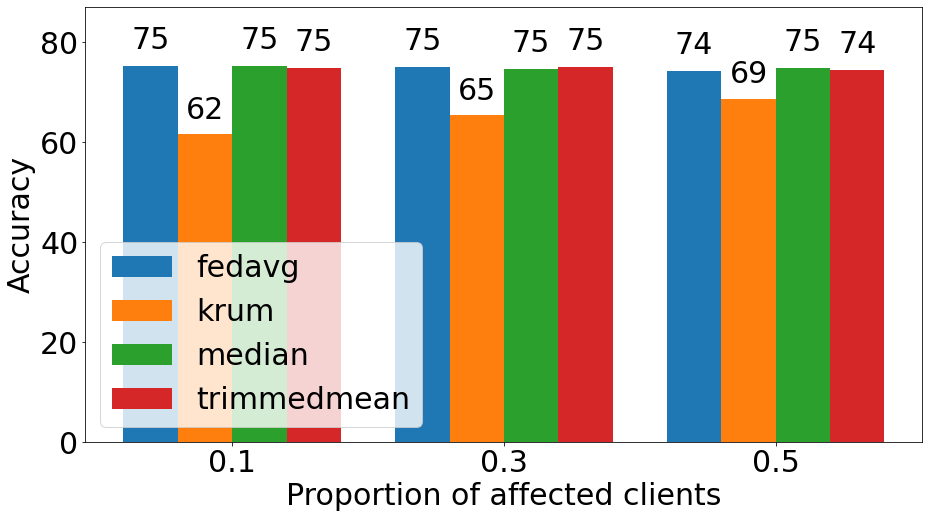}}
    \hspace{0.05\linewidth}
    \subfloat[Unbalance Mutator\label{adni_unbalance}]{%
        \includegraphics[width=0.40\linewidth]{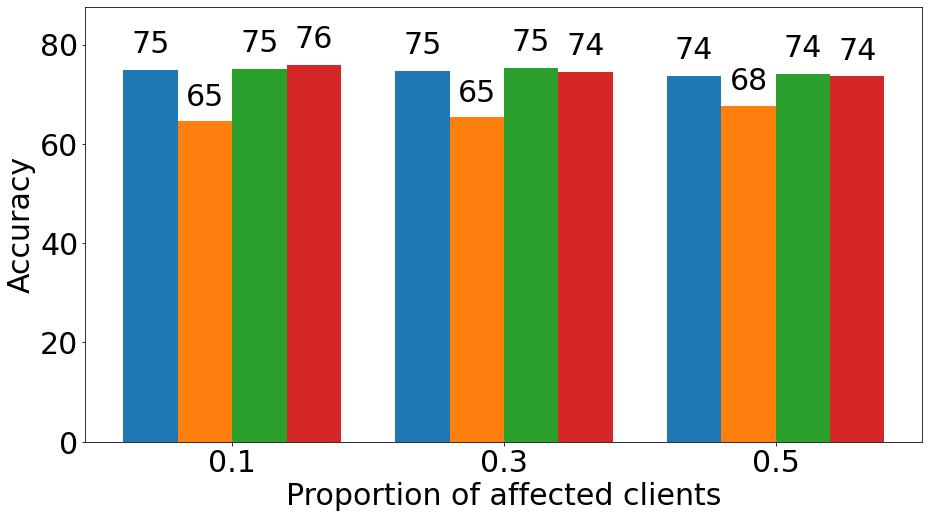}}
    \hfill
    \subfloat[Overlap Mutator\label{adni_overlap}]{%
        \includegraphics[width=0.40\linewidth]{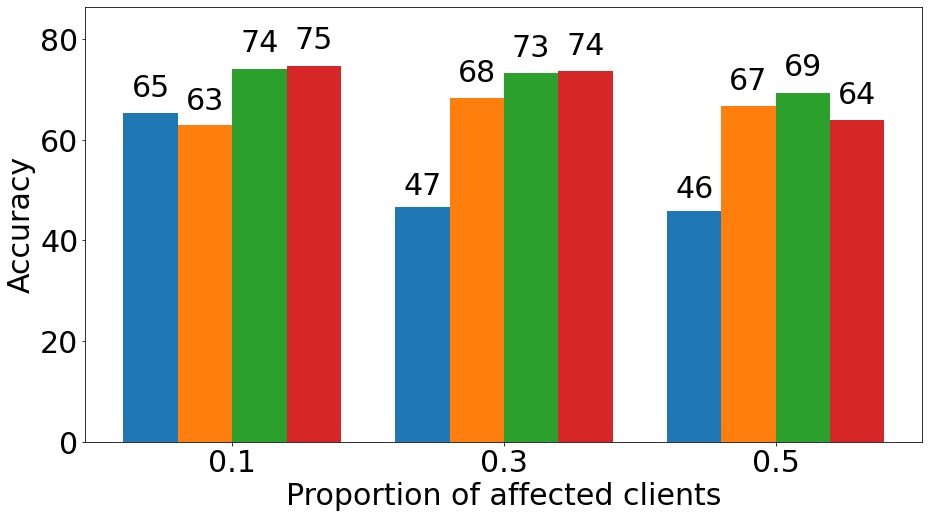}}
    \hspace{0.05\linewidth}
    \subfloat[Noise Mutator\label{adni_noise}]{%
        \includegraphics[width=0.40\linewidth]{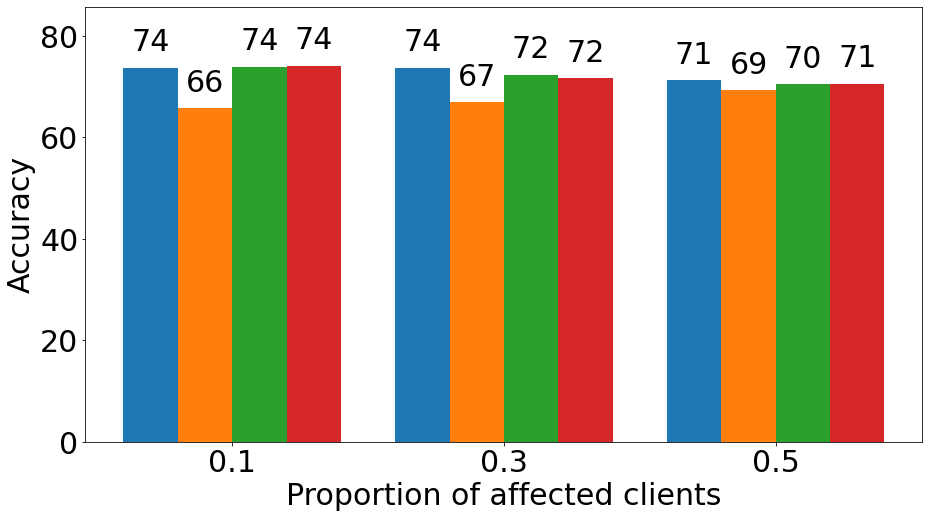}}
    \hfill
    \caption{ADNI - Aggregators performance under simulated faults.}
    \label{adni_aggregators_faults_fig}
\end{figure*}

\textbf{Effect of mutators on ADNI dataset:}

To see how aggregation methods work here and what the differences are compared to generic image datasets, we report the mutator results in Figure \ref{adni_aggregators_faults_fig}.

In most data mutators, the results are similar to the RQ1, and none of them significantly impact any of the aggregators. As seen before, the Noise mutator reduces the quality of the model slightly. However, interestingly, in the ADNI dataset, the Overlap mutator has a significant impact on Federated Averaging. As the proportion of affected clients increases, it performs even worse and hits 46\% accuracy. This can be because this dataset has only two classes, and this mutator overlaps only two classes (which are all of the classes here), as discussed in Section \ref{background_mutation}.

On the other hand, Median and Trimmed Mean show excellent robustness against this mutator, and in smaller proportions, they perform as if there was no attack. Although Median is slightly less resilient in smaller proportions, it performs better when half of the clients are byzantine. Krum is also more robust than Federated Averaging, but because of its problems with the ADNI dataset almost always performs worse than Median and Trimmed Mean.

An exciting conclusion is that robust aggregation methods can mitigate faults that hurt Federated Averaging, like when similar samples have different labels.

\begin{figure*}
    \centering
    \subfloat[Label Flip\label{adni_label}]{%
        \includegraphics[width=0.40\linewidth]{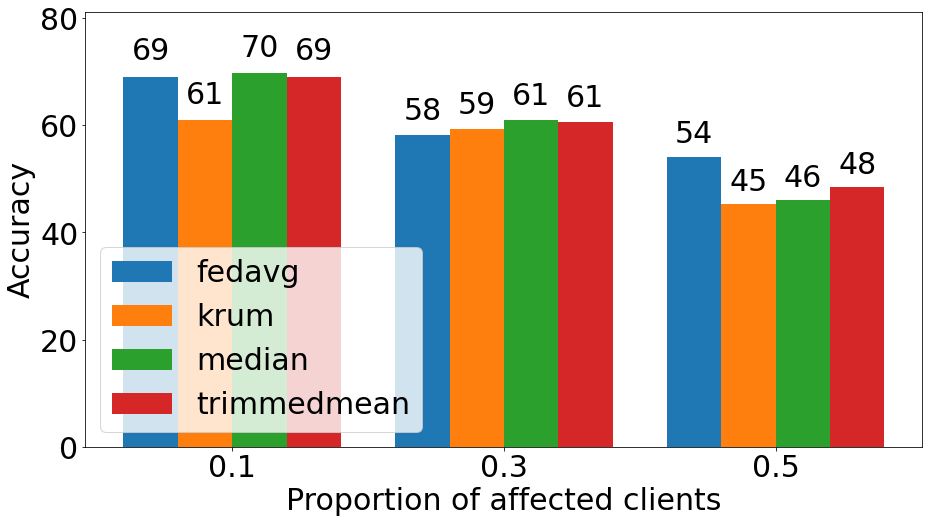}}
    \hspace{0.05\linewidth}
    \subfloat[Random Update\label{adni_random}]{%
        \includegraphics[width=0.40\linewidth]{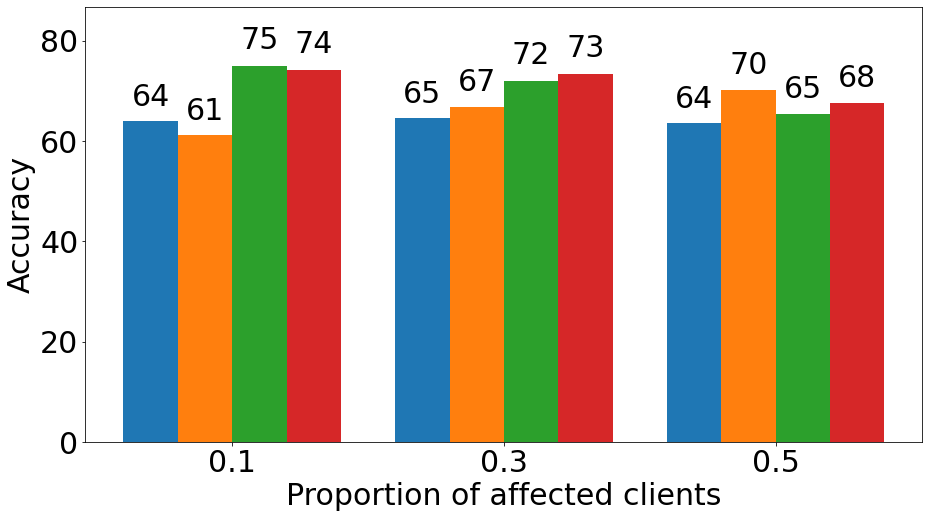}}
    \hfill
    \subfloat[Sign Flip\label{adni_sign}]{%
        \includegraphics[width=0.40\linewidth]{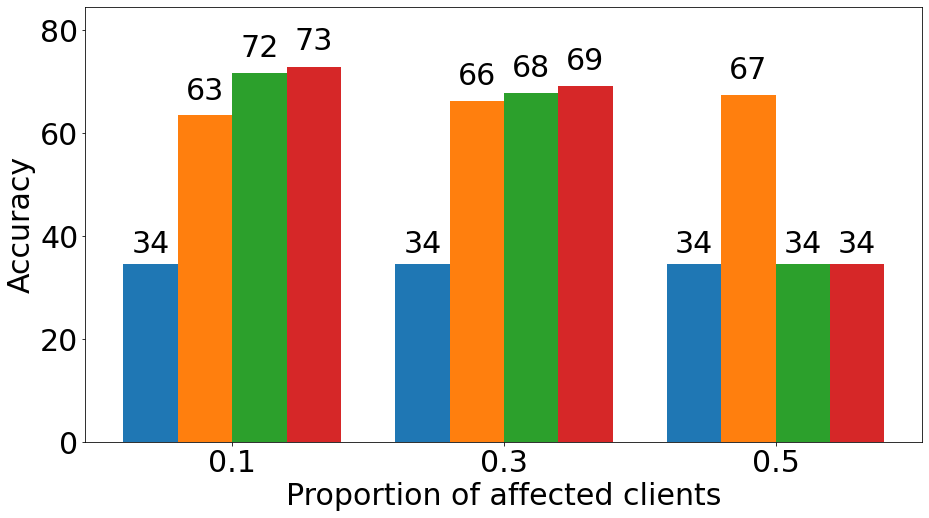}}
    \hfill
    \caption{ADNI - Aggregators performance under untargeted attacks.}
    \label{adni_aggregators_attacks_fig}
\end{figure*}

\textbf{Effect of attacks on ADNI dataset:}

According to Figure \ref{adni_label}, in the Label Flip attack, we see that in a small proportion of affected clients, all aggregators perform very similarly except Krum, which is worse, much like the previous RQ. As the proportion increases, we see that Median and Trimmed Mean get slightly better results. However, when half of the clients are under attack, they are less robust than Federated Averaging.

In untargeted model attacks (Figures \ref{adni_random} and \ref{adni_sign}), we see again that Federated Averaging is performing very poorly. In the Random Update, it achieves an average accuracy of 64\%. However, in the Sign Flip attack, it is even worse as models are guessing randomly and since the data is unbalanced final accuracy is 34\%. This shows that Sign Flip can cause more damage, and it makes sense because it is trying to guide the model in the opposite direction rather than a random direction.
In contrast, Trimmed Mean and Median are robust, and in the Random Update attack with smaller proportions, they even get close to no attack accuracy of 75\%. However, since Sign Flip is more powerful than Random Update, both aggregators get lower accuracy in the Sign Flip case and even completely fail when half of the clients are malicious.

Although Krum did not always converge to the best solutions, it is still more robust than Federated Averaging and is closely behind Median and Trimmed Mean. In some cases, like 0.5 proportion, it even works better than those two.

As a result, Trimmed Mean and Median for these attacks are the best and reasonably robust choices.
Furthermore, Median and Trimmed Mean aggregators show more robustness in higher proportions than RQ2, which can be because we are using transfer learning and many layers of the model are not trainable. Hence, model attacks are less effective than they were in RQ2.

\begin{figure*}
    \centering
        \includegraphics[width=0.40\linewidth]{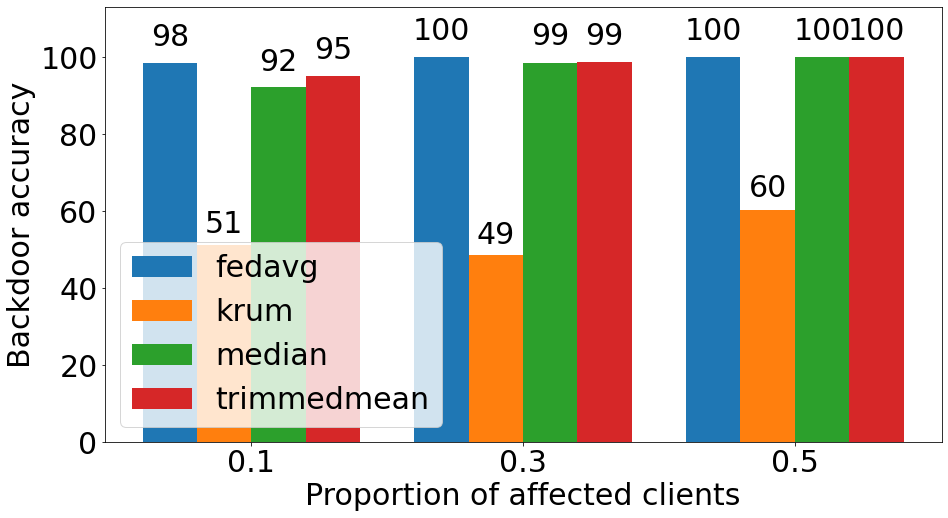}
    \caption{ADNI - Aggregators performance under Backdoor attack.}
    \label{adni_backdoor}
\end{figure*}

Figure \ref{adni_backdoor} shows the results for the Backdoor attack. The first observation is that Krum still is the most robust approach in the Backdoor attack with only 50\% backdoor task accuracy in small proportions. Even in the highest proportion, it still has 60\% backdoor accuracy, which is much better than the other aggregators. However, Krum struggles with the main task (as discussed before), so Krum has a trade-off between main and backdoor task accuracy.

Note that Krum's backdoor task accuracy in this dataset is much higher than in generic datasets used in RQ2 (more than 50\% in this question compared to less than 10\% in RQ2). This is because there are only two classes in this dataset, and the main task accuracy of the dataset does not exceed 75\% (for Krum, it is even lower). This means some samples are already misclassified to the attackers' target label, making the backdoor task more successful.

Moreover, the Median shows slight robustness in the smallest attack proportion against the backdoor objective, but it is inadequate. Trimmed Mean performs worse than Median, which is not good enough either. Federated Averaging is the worst of all, and its backdoor task is always successful in all cases. These patterns are much like what we observed in RQ2.

\begin{table}

    \caption{
    ADNI - Aggregators summary in terms of their accuracy (averaged across attacks and proportion of affected clients) and the number of times the aggregator is the best choice among all aggregators under study (Number of times achieving the Top rank) -- The shaded cells mark the best techniques.}
    \label{adni_summary}
    \centering
    
    \subfloat[Results for untargeted attacks and Overlap mutator (higher accuracy is better)]{
    \begin{tabular}{cccc}
    \bottomrule
        &      \multicolumn{2}{c}{Accuracy}& \multirow{2}{*}{\shortstack{Top rank\\frequency}}\\
         Aggregator & Mean & Std & \\
        \toprule
           FedAvg & 52.89 & 12.66 & 1\\
   Krum                & 63.23 & \cellcolor{gray!25}6.24 & 2\\
  Median              & 64.97 & 11.87 & 4\\
  Tri-mean        & \cellcolor{gray!25}65.16& 11.72 & \cellcolor{gray!25}5\\
    \bottomrule
    \end{tabular}}
    \hspace{.02\linewidth}
    \subfloat[Results for Backdoor attack (lower accuracy is better)]{
    \begin{tabular}{cccc}
    \bottomrule
        &      \multicolumn{2}{c}{Backdoor accuracy}& \multirow{2}{*}{\shortstack{Top rank\\frequency}}\\
         Aggregator & Mean & Std & \\
        \toprule
           FedAvg & 99.47 &\cellcolor{gray!25}0.75 & 0\\
   Krum                & \cellcolor{gray!25}53.34&5.05 & \cellcolor{gray!25}3\\
   Median              & 96.9 &3.41& 0\\
   Tri-mean        & 97.92&2.09 & 0\\
    \bottomrule
    \end{tabular}}
\end{table}

Following what we did in RQ2, we report the summary of aggregators for this dataset in Table \ref{adni_summary}. Unlike RQ2, the Overlap mutator is also included here since it showed to be effective for this dataset.

According to the table, Trimmed Mean gets the best results for untargeted attacks and mutators, and Median comes in second with a negligible difference. All aggregators lose the competition to the attacker for the Backdoor attack except Krum, which shows decent robustness, but its main task accuracy can still be problematic.

Lastly, we report the statistical tests. We have 4 attacks, 1 mutator that is effective(Overlap) proportions, and 3 proportions which result in 15 cases. Since Trimmed Mean and Median were close, we discuss their statistical test results in detail and compare them to the second-best aggregator (per case) in all the cases.

For Trimmed Mean, results show that it is significantly better in three cases. In five cases, it is better, but the difference is insignificant. In five cases, it is worse, but again, not significantly. Lastly, it is significantly worse in two cases. So in 86\% of cases, Trimmed Mean is the best choice. 

For Median, results show that it is significantly better in three cases. In four cases, it is better, but the difference is not significant. In six cases, it is worse, again not significantly. Finally, in two cases, it is significantly worse. So just like Trimmed Mean, in 86\% of cases, Median is the best choice. 
Also, this number for Krum and Federated Averaging is 60\% and 40\% respectively, which confirms that Median and Trimmed Mean are the best choices here.

\begin{tcolorbox}
\textbf{Answer to RQ3:}
Unlike RQ2, where Krum showed to be the best aggregator, overall, here, Trimmed Mean and Median are the most robust ones in 86\% of cases, and they can better maintain the quality of the FL process. However, Krum is still the best aggregator for the Backdoor attack (on the backdoor task, to be more precise). Moreover, Krum shows problems in the cross-silo FL with non-iid data, and it is not consistent like the other aggregators. Finally, even in a case-by-case situation in untargeted attacks, Trimmed Mean and Median are still a great choice of aggregation method.
\end{tcolorbox}

\subsubsection{\textbf{RQ4 results (an ensemble of aggregators):}
}
\label{results_rq4}





\begin{table*}

    \caption{Comparison of available aggregators with the ensemble aggregator.}
    \label{comb_summary}
    \centering
    \begin{tabular}{ccc|ccccc}
    \bottomrule
         &  &  & \multicolumn{5}{c}{Aggregator}\\
        Dataset & Attack & portion & FedAvg & Krum & Median & Tri-Mean & Ensemble\\
        \toprule
\multirow{8}{*}{CIFAR-10} & \multirow{3}{*}{Label Flip} & 0.1 & \cellcolor{gray!25}74.58 & 49.49 & 70.34 & 73.2 & 73.31\\
 &  & 0.3 & \cellcolor{gray!25}66.81 & 41.02 & 54.13 & 59.01 & 64.85\\
 &  & 0.5 & 36.4 & 29.41 & 27.59 & 31.84 & \cellcolor{gray!25}60.53\\
 \cline{2-8}
 & \multirow{3}{*}{Sign Flip} & 0.1 & 10.0 & 61.0 & 70.96 & 10.0 & \cellcolor{gray!25}72.08\\
 &  & 0.3 & 10.0 & \cellcolor{gray!25}60.62 & 10.0 & 10.0 & 59.77\\
 &  & 0.5 & 10.0 & \cellcolor{gray!25}59.13 & 10.0 & 10.0 & 10.0\\
 \cline{2-8}
 &\multicolumn{2}{c}{Average} & 34.63 & 50.11 & 40.5 & 32.34 & \cellcolor{gray!25}56.75\\
 &\multicolumn{2}{c}{Std} & 27.25& \cellcolor{gray!25}11.70 &25.92& 25.42 & 21.54\\
 \midrule
\multirow{8}{*}{ADNI} & \multirow{3}{*}{Label Flip} & 0.1 & 69.09 & 61.07 & 69.68 & 69.1 & \cellcolor{gray!25}70.03\\
 &  & 0.3 & 58.2 & 59.28 & 61.01 & 60.59 & \cellcolor{gray!25}65.46\\
 &  & 0.5 & 54.02 & 45.39 & 46.13 & 48.41 & \cellcolor{gray!25}65.83\\
 \cline{2-8}
 & \multirow{3}{*}{Sign Flip} & 0.1 & 34.49 & 63.46 & 71.61 & \cellcolor{gray!25}72.9 & 71.5\\
 &  & 0.3 & 34.49 & 66.17 & 67.75 & \cellcolor{gray!25}69.12 & 65.69\\
 &  & 0.5 & 34.49 & \cellcolor{gray!25}67.46 & 34.49 & 34.49 & 49.34\\
 \cline{2-8}
 &\multicolumn{2}{c}{Average}  & 47.46 & 60.47 & 58.45 & 59.1 & \cellcolor{gray!25}64.64\\
 &\multicolumn{2}{c}{Std}  & 13.73 & 7.30& 13.65& 13.62 & \cellcolor{gray!25}7.23\\
    \bottomrule
    \end{tabular}
\end{table*}

As discussed before, in this RQ, we evaluate our ensemble method on the CIFAR-10 (with non-iid=0.4) and ADNI datasets and test the new method on Label Flip and Sign Flip attacks.
We choose these attacks to have attacks from data and model poisoning categories. Furthermore, we select the Sign Flip attack over the Random update from the category of model poisoning attacks since it is more effective as it directs the training process in the opposite direction of the updates rather than just a random direction (we confirmed this in RQ2--3 results).
Given that the Fashion MNIST dataset is less challenging than CIFAR-10, based on RQ1--2 results, we only focus on these two datasets for the sake of space.

We report the results in Table \ref{comb_summary}. Out of these 12 configurations, the ensemble aggregator is the best option in Five cases, according to the median accuracies. Running statistical tests (Mann-Whitney U Test) shows that the ensemble is significantly better than the second technique in three cases out of five. However, in the remaining two cases, there is no significant difference. Furthermore, this difference is insignificant in four cases out of the seven remaining cases where the ensemble is not the best. As a result, in 75\% of cases, the ensemble aggregator is the most reasonable choice. If were run the same test for other aggregators, Federated Averaging gets 25\%, and the rest will achieve 33.3\%, which clearly shows the ensemble method is superior.
Lastly, our ensemble technique achieves the highest mean accuracy compared to other aggregators.

One caveat with our ensemble approach is that it takes longer to train. On average we see around 50\% increased training time which is not too bad considering the results. Furthermore, as the overhead is on the server, and servers are not limited like clients, servers can easily make up for this by utilizing more GPU resources. Lastly, this method only has an overhead in the training phase, so clients will not notice any difference in the testing phase (this phase is repetitive and is where users interact with the model).

\begin{tcolorbox}
    \textbf{Answer to RQ4:}
     It is possible to offer an ensemble aggregator that is more generalizable (the best choice in 75\% of the cases) than any of its constituent aggregators. Also, it shows decent robustness in both data (Label Flip was a representative of this category) and model poisoning (Sign Flip was a representative of this category) attacks without prior knowledge.
\end{tcolorbox}


\subsection{Discussions}

As discussed, one of the main challenges that can jeopardize the quality of the FL is byzantine attacks. Our results in Section \ref{results_rq1}  confirm this problem and show that attacks and, to some extent, faults can degrade the overall quality of the FL. In Section \ref{results_rq2} we saw that having robust aggregation techniques instead of the basic Federated Averaging can indeed improve the robustness and, consequently, the quality of training.
Although robust aggregators are effective, another factor that challenges the quality of the FL process is the proportion of byzantine clients as shown in Section \ref{results_rq2}. The aggregators typically have a breaking point regarding the number of byzantine clients, and the aggregator might not work past that as intended. For instance, Median's breaking point is when half of the clients are byzantine \citep{lyu2020privacy}. However, the number of byzantine clients might not be below the aggregators' breaking point in a real scenario. As a result, there could still be serious quality issues in practice.

Moreover, the choice of the aggregator is non-trivial, and it is one of the most significant challenges we currently face in this domain. Our experiments in Sections \ref{results_rq2} and \ref{results_rq3} show that there is no perfect aggregator for all cases, and it is imperative to consider the context and settings when applying a specific aggregator.

Even though selecting aggregators based on the settings seems like a reasonable tactic, it is usually not practical. Since in a real case, essential factors like the dataset and attack type are unknown to the server (consequently the aggregator), i.e., there is no way to know which aggregator would work best. However, in Section \ref{results_rq4} results confirm that an ensemble of all aggregators can indeed result in a better and more general aggregator.

Moreover, as discussed in Section \ref{threat_model}, some aggregators need extra information to perform decently, which might not be available. For instance, Krum and Trimmed Mean need to know how many byzantine clients are at each round, which is not a realistic assumption. So, using them in an actual application would be problematic unless they use some estimation technique to estimate the number of byzantine clients. Although this seems a solid solution, the choice of estimation technique itself is significant and needs extensive study.

An important point to notice is that model attacks, which are far more powerful than poisoning attacks (Shown in Section \ref{results_rq1}), are only possible if the attackers can access and alter the client code. Consequently, it would be ideal to improve the client code security and stop these attacks before they happen. So it would be interesting to see which techniques can be applied here to make it harder for the attacker to implement model poisoning attacks. One idea would be cryptography which is already being used on secure aggregators \citep{secagg}.

Another exciting follow-up research opportunity in this domain is the client selection strategy in a cross-device FL. The baseline for client selection is choosing clients randomly, at each round \citep{mcmahan2017communicationefficient}. Recently, new client selection techniques have been introduced to make the overall FL process faster by considering devices' hardware heterogeneity \citep{fedcs, tifl, oort}. It would be interesting to see how a byzantine-aware client selection technique could improve the overall quality alone and how it would work with a robust aggregation technique.


\subsection{Limitations and threats to validity}
One of the limitations of our study is that in RQ4, our proposed technique is not applicable in the Backdoor attack where the backdoor task is unknown. We will consider this as future work and an extension of this study.

In terms of internal validity, we reused well-known attacks and defenses (robust aggregations) from the existing literature and their replication packages. Two co-authors of the papers carefully went through the implementations. Furthermore, an important point is that we did not use FL-specific frameworks like Tensorflow Federated (TFF). The reason is that many of the aggregators used in our study are not available in FL frameworks. Also, integrating them into the frameworks is not a straightforward task, and it is impossible in some cases due to restricted APIs (e.g., implementation of Krum in TFF).
Furthermore, client-server communications are actually very important and critical features, when one wants to deploy FL. However, in our context and to compare the attacks and defenses, a single machine simulation is enough and the deployment communication features will not affect the aggregator or attacks.
The last question that may arise here is whether a simulation approach like ours is going to have the same performance (in terms of the model accuracy) as a framework with a distributed deployment. The answer is that if factors like the DL framework and its version, the python version, and the hardware are identical and communications are reliable, the results should be the same.

Concerning external validity, although we have broad systematic coverage of many existing approaches, the choice of dataset and models and the number of attacks, mutators, and aggregators can be questioned. 
To mitigate this problem, regarding datasets, we used well-known generic Fashion MNIST and CIFAR-10 datasets and an application-specific ADNI dataset, which is a real-world federated dataset, to improve the generalizability of the findings. 
In terms of the models, although like many related works  \citep{rsapaper,ditto} and to make the systematic experiments with all combinations manageable, we used only one model per dataset, we used a diverse set of models across datasets (from simple CNNs to more complex VGG16).
We tried to pick well-established FL untargeted and targeted attacks from both data and model categories regarding the selected attacks. Additionally, we used previously studied mutation operators that were applicable to FL and were based on real faults. 
Furthermore, concerning the aggregators, we studied four of the most well-studied techniques in the published FL literature that come with replication packages.
Finally, a question may arise regarding the ensemble method used in RQ4, like why is the technique not considering identifying and filtering the byzantine clients instead? The answer is that aggregators like Krum are doing the same thing, and the proposed method should include a new technique to make up for instances where single aggregators do not perform as expected. Thus, an ensemble will be helpful, and that is why we chose it.

Regarding construct validity, we used prediction accuracy as the metric for all the untargeted attacks and faults. Also, for the Backdoor attack, we used the backdoor task accuracy to be consistent with related studies.

In terms of conclusion validity, one potential threat could be the random factors of the study, like client selection for training and selection of byzantine clients. We tried to mitigate this issue by running the experiments 10 times and reporting the median of the results to exclude the outliers. Also, we ran statistical significance tests to show that the observations were not due to chance.

\section{Related work}
\subsection{Robust Federated Learning:} 
In recent years, many techniques have been proposed to increase the robustness of FL against the byzantine attacks. These techniques fall into two main categories that we will discuss.

\textbf{Aggregation techniques:}
The first category is aggregation techniques which include the techniques used in our study like Krum, Median, and Trimmed-Mean \citep{krumpaper,medianpaper}. This is the category that is the focus of this study.

However, other techniques have been proposed like RSA that ensures the robustness in heterogeneous cases (non-iid data)  \citep{rsapaper}. RSA is different from other robust aggregations discussed in this paper as it is not comparing updates together. Instead, it compares the received models (not the updates) at the server with the global model. It is a norm-based approach that penalizes models that deviate too much from the global model by limiting their contribution to the aggregated model. We could not use this technique since its incomplete replication package was not well-documented.

Another technique introduced recently is called attack-adaptive aggregation, which works with attention models  \citep{adaptive}. The objective of the proposed aggregation model is to assign a weight to each local update in a way that byzantine updates get excluded (they get zero weight assigned to them) and do a weighted averaging at the end. The main drawback of this attention-based aggregation model is that it first needs to be trained on the data collected during a normal FL process (the data is the local updates sent to the server). After the model is trained, it can be used in another FL process for aggregation. Moreover, this technique is neither scalable nor generalizable to different cases. Since our study was comprehensive and this adaptive technique was not generalizable and had an incomplete replication package,  we excluded this technique from our experiments.

Furthermore, another aggregation has been proposed called SEAR, which works in secure environments \citep{sear}. Since secure aggregation uses cryptography, it stops the server from seeing the updates and makes these robust aggregation techniques impossible to use. They proposed a new secure aggregation called SEAR, which relies on Intel Software Guard Extensions (SGX) for secure aggregation. They also proposed a new aggregation technique that works best with the SGX environment. However, their final results regarding final model accuracy are not that different from the aggregators studied in this paper. Since their main contribution is to the security and performance of robust aggregators in the secure environment (and their aggregator alone is not that powerful), we omitted SEAR from our experiments.

There are also techniques proposed online \citep{residual, rfa} which are not published in any peer-reviewed venue yet. Thus we have not included them in this study.

\textbf{Optimization techniques:}
Other than aggregation techniques, new optimization techniques have been proposed to improve the overall quality of FL. The first and most popular technique in this category is FedProx  \citep{fedprox}. The main idea of FedProx is to change the learning rates of clients based on their data distribution to make the convergence faster. FedProx is not designed to resist attacks. Instead, its goal is to improve Federated averaging in non-iid cases.
However, a similar idea has been used in another work called FoolsGold to make FL more robust against Sybil attacks.  \citep{foolsgold}. Their approach works based on model updates received at the server. FoolsGold first compares the update vectors based on their similarity and identifies the problematic updates and, consequently, the byzantine clients. Afterward, FoolsGold changes the learning rate of the byzantine clients to make their updates have less contribution in the final model. However, this approach does not work in several cases, like model poisoning attacks. If the attacker sends random updates, the learning has no effect in the final update, and in other cases, clients can multiply their updates without restriction to make up for the learning rate. Furthermore, in model poisoning attacks, the attacker has full control of the byzantine clients; thus, it can set its learning rate to make its attack more effective \citep{howtoback}, nullifying what FoolsGold was trying to achieve. 

More recently, a new scheme has been proposed called Ditto  \citep{ditto}. Unlike previous techniques that only change the learning rate, Ditto introduces a completely new optimization algorithm that follows a particular objective to ensure fairness and robustness. 

Since these optimization techniques were not the main focus of our study, we did not include them in our experiments.

\subsection{Empirical evaluation of FL attacks and defenses} 

The effect of attacks and aggregators on FL quality is an important topic, and it has been studied extensively. 
\citeauthor{localpoison} conducted a study on FL aggregators and attacks. They used Label Flip, Random Update, and their proposed attack with Krum, Median, and Trimmed Mean aggregators  \citep{localpoison}. They also have studied the effects of non-iid distribution and the proportion of affected clients but only for one dataset. However, our study has some key differences from their study. Firstly, they omitted Federated Averaging, the baseline of FL, and as we showed before, is better than the other aggregators in the Label Flip attack. Secondly, they only used untargeted attacks in their evaluations, whereas we also studied targeted attacks and mutators to simulate faults. Lastly, they only used naturally centralized datasets in a cross-device setting; in contrast, we also did a comprehensive study on a federated dataset and cross-silo FL setting.

Like the previous related study, \citeauthor{advlens} studied attacks in FL (more specifically model poisoning) and introduced a new attack technique \citep{advlens}. They introduced a targeted model poisoning attack and showed its effectiveness against Federated Averaging on the Fashion MNIST dataset. They also compared Krum and Median using their attack. This study is more focused on a variation of the targeted attacks. In contrast, we conducted a large-scale study on both untargeted and targeted attacks, mutators, and more aggregation techniques on all attacks. We also considered three datasets, one of which was naturally federated, and studied the effect of different distributions and the proportion of attacked clients.

Another work has surveyed attacks and defenses in FL, which is more theoretical than the previously discussed studies  \citep{lyu2020privacy}. They studied the attacks and defenses and compared them using their theoretical details and limitations. Furthermore, they studied different privacy techniques used in FL. Consequently, our empirical study adds a practical value to their theoretical study and explores mentioned limitations in practice.

A summary of this paper's contributions compared to some of the more related works is reported in Table \ref{related_summary}.

\begin{table}

    \caption{Contributions of this study compared to related works.}
    \label{related_summary}
    \centering
    \begin{tabular}{>{\centering\arraybackslash}m{0.14\linewidth}>{\centering\arraybackslash}m{0.1\linewidth}>{\centering\arraybackslash}m{0.1\linewidth}>{\centering\arraybackslash}m{0.1\linewidth}>{\centering\arraybackslash}m{0.1\linewidth}>{\centering\arraybackslash}m{0.1\linewidth}>{\centering\arraybackslash}m{0.1\linewidth}}
    \bottomrule
    &\cite{rsapaper}&\cite{adaptive}&\cite{localpoison}&\cite{advlens}&\cite{lyu2020privacy}& Our study\\
    \toprule
        Aggregators &\checkmark &\checkmark & \checkmark&\checkmark & \checkmark& \checkmark\\       
        \midrule
        \shortstack{Untargeted\\attacks} &\checkmark &\checkmark &\checkmark & \checkmark&\checkmark & \checkmark \\        \midrule
        \shortstack{Targeted\\attacks} & & \checkmark& & \checkmark& &  \checkmark\\        \midrule
        Faults & & & & & & \checkmark \\   
        \midrule
        Cross-device &\checkmark & \checkmark& \checkmark&\checkmark &\checkmark & \checkmark \\  
        \midrule
        Cross-silo & & & & & &  \checkmark\\  
        \midrule
        Multiple distributions &\checkmark& & \checkmark& & \checkmark&  \checkmark\\
        \midrule
        \shortstack{New aggregation\\technique} & \checkmark&\checkmark & & & & \checkmark \\ 
        \midrule
        Limitations & Technique needs tuning for different cases & Not generaizable and scalable& Federated Averaging is not considered &Does not consider different proportion of attackers & The study is purely theoretical & New technique works only on untargeted attacks \\      
    \bottomrule
    \end{tabular}
\end{table}

\section{Conclusion and future work}

In this paper, we conducted a large-scale empirical study on the effect of faults and attacks on FL aggregators. We performed our experiments on two generic image datasets, each with three different distributions, one federated medical dataset, eight attacks and mutators, and four aggregation techniques resulting in 496 configurations.
Results show that the Sign Flip and Backdoor attacks are the most effective attacks. Moreover, mutators do not significantly impact FL's quality, except for the Overlap data mutator, which can affect Federated Averaging in the ADNI dataset. In addition, our study shows that there is no single best robust aggregator, and their accuracy depends on factors such as attack type, dataset, and data distribution. For instance, Krum is most robust in model poisoning attacks, but it is not acceptable in the Label Flip attacks. Inspired by the results of different aggregators, we show that an ensemble of these aggregators can be more robust than (or as good as) any single aggregator to improve the FL process quality in 75\% of cases where the attacks and data distribution are unknown to the aggregator.
In the future, we plan to study different attack scenarios, e.g., a case where the attacker is aware of the aggregator used on the server. Also, we want to extend this study to other FL-exclusive issues, such as clients' machines capabilities, learning frameworks, network failures, and arithmetic computation precision issues. Lastly, we want to extend the ensemble technique to make it effective against targeted attacks like the Backdoor attack and more efficient by using different heuristics while selecting the best aggregator in each round.

\begin{acknowledgements}
Data collection and sharing for this project were funded by the Alzheimer's Disease Neuroimaging Initiative (ADNI) (National Institutes of Health Grant U01 AG024904) and DOD ADNI (Department of Defense award number W81XWH-12-2-0012). ADNI is funded by the National Institute on Aging, the National Institute of Biomedical Imaging and Bioengineering, and through generous contributions.
\end{acknowledgements}

\bibliographystyle{spbasic}      
\bibliography{refs}

\end{document}